\documentclass{article}

\usepackage[preprint, nonatbib]{neurips_2023}
\usepackage{bm}
\usepackage{bbm}
\usepackage[square, sort, numbers]{natbib}
\setcitestyle{numbers}

\usepackage[dvipsnames]{xcolor}

\usepackage[utf8]{inputenc} 
\usepackage[T1]{fontenc}    
\usepackage[colorlinks=true]{hyperref}
\hypersetup{
    colorlinks,
    citecolor=Maroon,
    filecolor=Maroon,
    linkcolor=Maroon,
    urlcolor=Maroon
}
\usepackage{url}            
\usepackage{booktabs}       
\usepackage{amsfonts}       
\usepackage{amsmath}
\usepackage{nicefrac}       
\usepackage{microtype}      
\usepackage{xcolor}         
\usepackage{graphicx}
\usepackage[capitalise]{cleveref}
\graphicspath{{./figures/}}
\usepackage{algorithm}
\usepackage{multirow}
\usepackage[dvipsnames]{xcolor}
\usepackage{listings}
\usepackage{soul}

\usepackage{tcolorbox}

\definecolor{codegreen}{rgb}{0,0.6,0}
\definecolor{codegray}{rgb}{0.5,0.5,0.5}
\definecolor{codepurple}{rgb}{0.58,0,0.82}
\definecolor{backcolour}{rgb}{0.97,0.97,0.95}

\lstdefinestyle{mystyle}{
    backgroundcolor=\color{backcolour},   
    commentstyle=\color{codegreen},
    keywordstyle=\color{magenta},
    numberstyle=\tiny\color{codegray},
    stringstyle=\color{codepurple},
    basicstyle=\ttfamily\footnotesize,
    breakatwhitespace=false,         
    breaklines=true,                 
    captionpos=b,                    
    keepspaces=true,                 
    numbers=left,                    
    numbersep=5pt,                  
    showspaces=false,                
    showstringspaces=false,
    showtabs=false,                  
    tabsize=2
}

\newcommand\YAMLcolonstyle{\color{red}\mdseries}
\newcommand\YAMLkeystyle{\color{black}\bfseries}
\newcommand\YAMLvaluestyle{\color{blue}\mdseries}
\makeatletter

\newcommand\language@yaml{yaml}
\expandafter\expandafter\expandafter\lstdefinelanguage
\expandafter{\language@yaml}
{
    escapeinside={(*@}{@*)},
    backgroundcolor=\color{backcolour},
    keywords={true,false,null,y,n},
    keywordstyle=\color{darkgray},
    basicstyle=\YAMLkeystyle,                                 
    sensitive=false,
    comment=[l]{\#},
    morecomment=[s]{/*}{*/},
    commentstyle=\color{purple}\ttfamily,
    stringstyle=\YAMLvaluestyle\ttfamily,
    moredelim=[l][\color{orange}]{\&},
    moredelim=[l][\color{magenta}]{*},
    moredelim=**[il][\YAMLcolonstyle{:}\YAMLvaluestyle]{:},   
    morestring=[b]',
    morestring=[b]",
    literate =    {---}{{\ProcessThreeDashes}}3
                {>}{{\textcolor{red}\textgreater}}1     
                {|}{{\textcolor{red}\textbar}}1 
                {\ -\ }{{\mdseries\ -\ }}3,
    numberstyle=\tiny\color{codegray},
    numbers=left,                    
    numbersep=5pt,
    basicstyle=\ttfamily\footnotesize,
}

\lst@AddToHook{EveryLine}{\ifx\lst@language\language@yaml\YAMLkeystyle\fi}
\makeatother

\newcommand\ProcessThreeDashes{\llap{\color{cyan}\mdseries-{-}-}}

\usepackage[noend]{algpseudocode}
\title{Pangu-Agent: A Fine-Tunable Generalist Agent\\with Structured Reasoning}

\author{%
  Filippos Christianos~\textsuperscript{\textnormal{1}}\thanks{Contributing equally to this work.} 
  \And
  Georgios Papoudakis~\textsuperscript{\textnormal{1}}\footnotemark[1] 
  \And
  Matthieu Zimmer\textsuperscript{\textnormal{1}}
  \And
  Thomas Coste\textsuperscript{\textnormal{1}}
  \And
  Zhihao Wu\textsuperscript{\textnormal{1}}
  \And
  Jingxuan Chen\textsuperscript{\textnormal{1}}
  \And
  Khyati Khandelwal\textsuperscript{\textnormal{1}}
  \And
  James Doran\textsuperscript{\textnormal{1}}
  \And
  Xidong Feng\textsuperscript{\textnormal{2}}\thanks{Work done during an internship at Huawei.}
  \And
  Jiacheng Liu\textsuperscript{\textnormal{2}}\footnotemark[2]
  \And \And \And 
  Zheng Xiong\textsuperscript{\textnormal{3}}\footnotemark[2]
  \And   \And \And 
  Yicheng Luo\textsuperscript{\textnormal{2}}\footnotemark[2]
  \And   \And \And 
   Jianye Hao\textsuperscript{\textnormal{1}} \\ 
 \And \And 
  Kun Shao\textsuperscript{\textnormal{1}}\footnotemark[3]
  \And
  Haitham Bou-Ammar\textsuperscript{\textnormal{1,2}}\thanks{Corresponding authors: \{shaokun2, haitham.ammar\}@huawei.com, jun.wang@cs.ucl.ac.uk.}
  \And
  Jun Wang\textsuperscript{\textnormal{2}}\footnotemark[3] 
  \And
  \textnormal{\hspace{-0.15cm}\textsuperscript{1}Huawei Noah’s Ark Lab,
  \textsuperscript{2}University College London,
  \textsuperscript{3}University of Oxford
  }
}
\date{
  \textsuperscript{1}Huawei Noah’s Ark Lab,\\
  \textsuperscript{2}University College London,\\
  \textsuperscript{3}University of Oxford.\\
//  \textsuperscript{4}Tianjin University\\
  }

\newtoggle{final}
\toggletrue{final}
\togglefalse{final} 

\DeclareMathOperator{\mem}{\mathtt{MEM}}
\DeclareMathOperator{\llm}{\mathtt{LLM}}

\DeclareMathOperator*{\E}{\mathbb{E}}

\begin{document}

\maketitle

\begin{abstract}
  A key method for creating Artificial Intelligence (AI) agents is Reinforcement Learning (RL). However, constructing a standalone RL policy that maps perception to action directly encounters severe problems, chief among them being its lack of generality across multiple tasks and the need for a large amount of training data. The leading cause is that it cannot effectively integrate prior information into the perception-action cycle when devising the policy. Large language models (LLMs) emerged as a fundamental way to incorporate cross-domain knowledge into AI agents but lack crucial learning and adaptation toward specific decision problems. This paper presents a general framework model for integrating and learning structured reasoning into AI agents' policies. Our methodology is motivated by the modularity found in the human brain. The framework utilises the construction of intrinsic and extrinsic functions to add previous understandings of reasoning structures. It also provides the adaptive ability to learn models inside every module or function, consistent with the modular structure of cognitive processes. We describe the framework in-depth and compare it with other AI pipelines and existing frameworks. The paper explores practical applications, covering experiments that show the effectiveness of our method. Our results indicate that AI agents perform and adapt far better when organised reasoning and prior knowledge are embedded. This opens the door to more resilient and general AI agent systems.
\end{abstract}

\section{Introduction}\label{sec:introduction}
\setcounter{footnote}{0} 

Since the inception of AI, developing generalist agents capable of solving and adapting to increasingly complex tasks has been an overarching goal. Such agents can be defined as: \emph{"... entities that appropriately act in uncertain environments, where appropriate actions increase the probability of success and where 
success is the achievement of behavioural subgoals that support the system's ultimate goal." (J. S. Albus~\citep{albus1991outline}).} Hence, AI agents must be equipped with sensors to perceive environments and methods to act upon them to achieve their goals. This definition can capture diverse autonomous systems, ranging from simple software programs running on the web to complex embodied entities interacting in unknown and stochastic environments. 

AI agents, vital to many intelligent applications, are often trained through RL, which develops decision-making skills through environmental interactions. Both model-based and model-free deep RL methodologies have shown significant achievements, such as AlphaZero's board game proficiency~\citep{silver2018general}, improved sorting and multiplication algorithms~\citep{mankowitz2023faster,fawzi2022discovering}, drone racing championships~\citep{kaufmann2023champion} and plasma control in fusion reactors \citep{degrave2022magnetic}. 
These successes involved a standard RL pipeline, where agents learn what we refer to as an \emph{extrinsic function} -- a policy that directly interacts with the outside world, i.e., responding to environmental stimuli to maximise a reward signal. This function, often a parameterised neural network, generates actions based on environmental observations.

The classic RL approach, using a single mapping function to define a policy \(\pi\), often proves inadequate in complex environments, contradicting the aim of generalist agents to interact, adapt, and learn across multiple stochastic environments. This sparked studies on improved learnability via meta-RL \citep{Rakelly2019EfficientOM,Gupta2018MetaReinforcementLO,Wang2018PrefrontalCA,Nagabandi2018LearningTA}, intrinsic motivation-based exploration \citep{Barto2013IntrinsicMA, Tang2016ExplorationAS, Burda2018ExplorationBR}, auxiliary tasks \citep{Jaderberg2016ReinforcementLW,Shelhamer2016LossII,li2019hierarchical}, inverse RL \citep{Ng1999PolicyIU, Devlin2012DynamicPR,Goyal2019UsingNL}, Bayesian priors \citep{Brochu2010ATO,Poupart2006AnAS,ghavamzadeh2015bayesian}, and hierarchical policies \citep{Kulkarni2016HierarchicalDR,barto2003recent,nachum2018data,HRL}. However, RL-introduced priors are often task-specific and require extensive engineering and domain expertise. To allow for more general priors, recent studies have shifted towards integrating large language models (LLMs) into agent frameworks, as seen in AutoGen~\citep{wu2023autogen}, AutoGPT~\citep{gravitas2023auto}, and AgentVerse~\citep{chen2023agentverse}. Research on LLM-based AI agents not only uses LLMs as foundational priors but also incorporates tools~\citep{schick2023toolformer,shen2023hugginggpt} and multi-agent communication~\citep{chen2023agentverse} to build generalist agents. Existing frameworks, outlined in \cref{tab:agent_summary} and discussed in \cref{sec:related}, assume fixed reasoning structures and typically lack fine-tuning for new behaviours and skills, limiting agents to the pre-trained LLM's capabilities and posing a risk in out-of-distribution tasks.

\begin{figure}[t!]
    \centering
    \includegraphics[trim={0em 15em 0em 26em },clip,width=\textwidth]{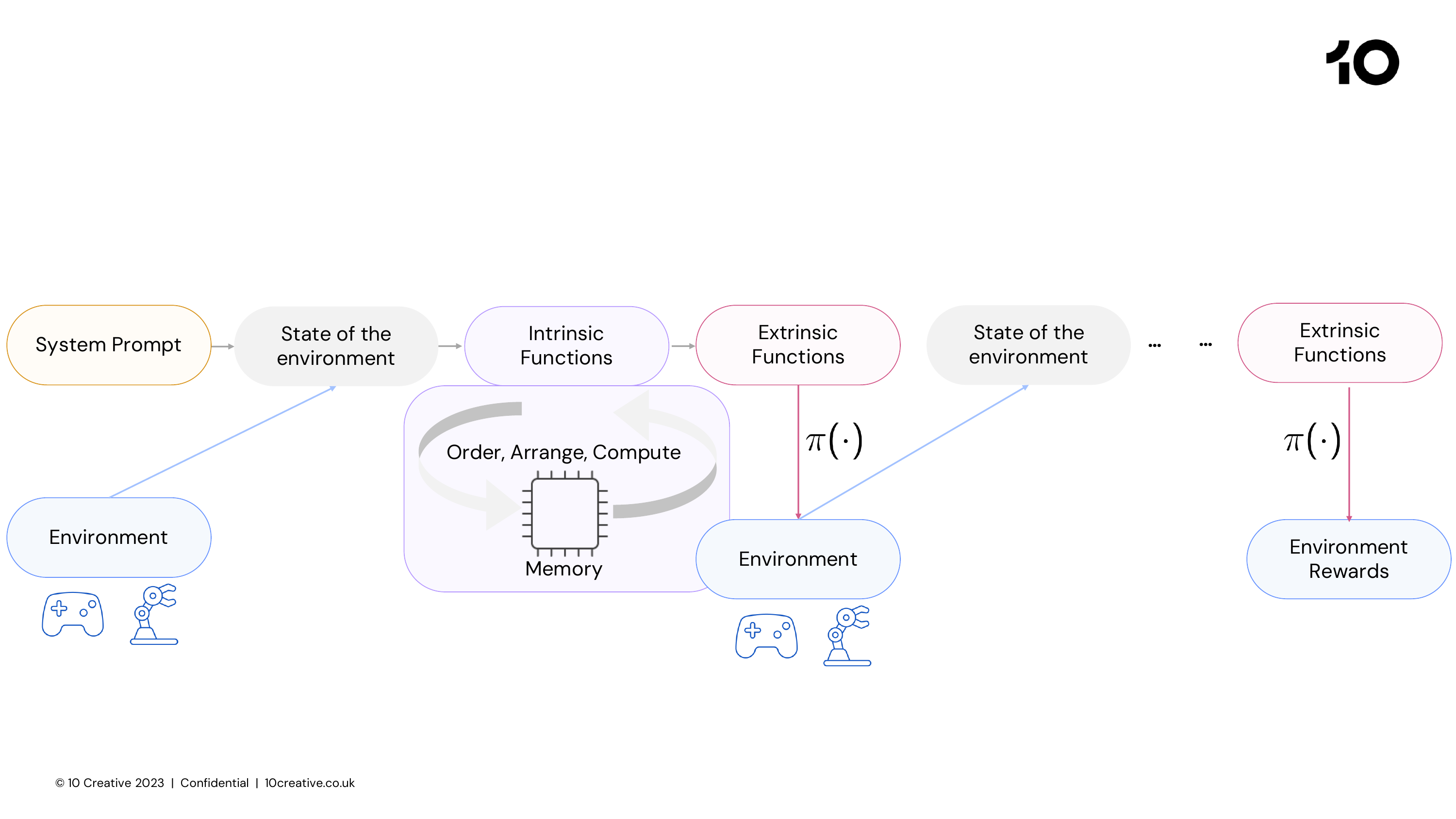}
    \caption{Pictorial depiction of the Pangu-Agent pipeline with RL. Starting from the system prompt and initial state, our agent executes actions in the environment and observes the next state and the reward. The trajectories generated can be used for finetuning the LLM.}
    \label{Fig:RL}
\end{figure}

\begin{figure}[t!]
    \centering
    \includegraphics[trim={0em 26em 0em 26em },clip,width=.9\textwidth]{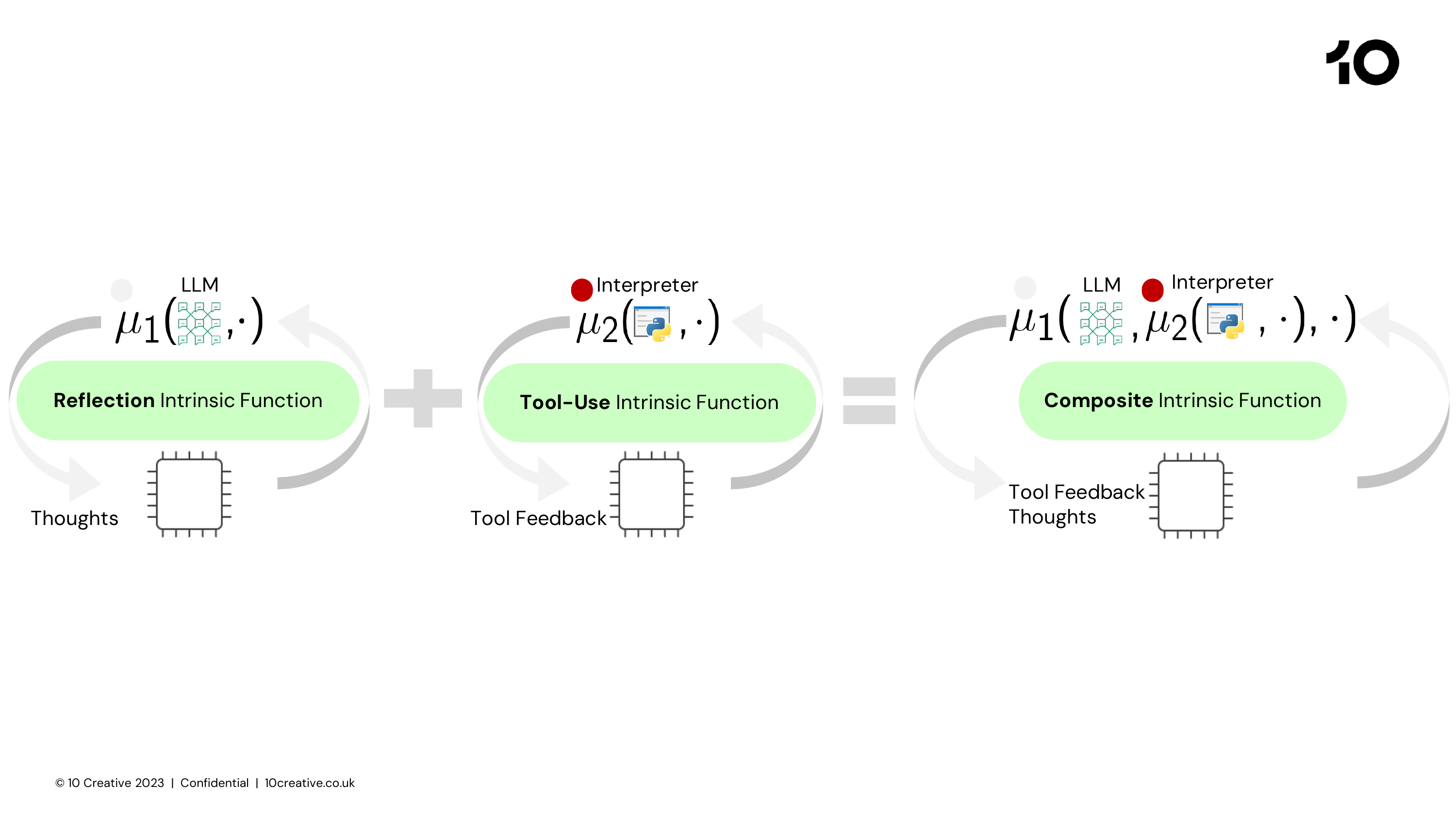}
    \caption{Visualisation of three intrinsic functions demonstrating our formulation's importance in improving our agent's modularity and flexibility. The intrinsic functions can be re-defined and re-configured by users, e.g., $\mu_{1}(\cdot)$ taking an LLM as input to produce thoughts or $\mu_2 (\cdot)$ utilising tools to help improve reasoning. We also support nesting those intrinsic functions to build more general modules for complex and challenging decision-making tasks. }
    \label{fig:pangu_visual0}
\end{figure}

In this paper, we propose the \emph{Pangu-Agent} framework 
 (also see \cref{Fig:RL})  as a first step towards remedying the above two problems. Our work distinguishes itself from prior frameworks in two critical ways:
\textit{i)} we formalise the internal thinking process of the agent as a form of \emph{structured reasoning} and 
\textit{ii)} we show how agents can be fine-tuned via supervised learning and RL. To introduce structured reasoning, we assume a family of \emph{intrinsic functions} $\bm{\mu}(\cdot)$ 
that act on and transform the agent's internal memory.
Introducing these intrinsic functions can reformulate the typical RL objective to one that supports multiple `thinking' steps. 
Therefore, the typical RL objective which aims to find a policy \(\pi\) conditioned on the history of observations \(\Vec{\bm{o}}\) that maximises the returns \(\textbf{R}\), i.e.\ \(\max_{\pi(\cdot)} \textbf{R}(\pi(\cdot|\Vec{\bm{o}}))\) can be rewritten using a nested set (see \cref{fig:pangu_visual0}) of intrinsic functions \(\Vec{\bm{\mu}}(\cdot)\) as:
\begin{equation}
\label{Eq:NewFormulation}
    \overbrace{\max_{\pi(\cdot)} \textbf{R}(\pi(\cdot|\Vec{\bm{o}}))}^{\text{Standard RL}} \rightarrow \overbrace{\max_{\pi(\cdot),\vec{\bm{\mu}}(\cdot)} \textbf{R}(\pi(\cdot|\Vec{\bm{o}}, \Vec{\bm{\mu}}(\Vec{\bm{o}}))).}^{\text{Pangu Opt.}}
\end{equation}
These nested intrinsic functions have been absent in RL formulations, whereby standard RL focuses on directly learning policies that output actions from perceptions. While it is customary to parameterise policies by deep network architectures, we contend that the absence of inherent reasoning structures in standard RL pipelines can become a significant bottleneck when scaling agents across tasks via foundational model policies precisely because gradients fail to offer enough supervision for all deep layers with only (sparse) rewards as guidance. Our framework demonstrates how structured reasoning can empower RL to surmount these challenges, leveraging large-scale foundational models to provide priors and enable generalisation across broad domains.
\begin{table}[t]
\label{Tab:Diff}
  \centering
  \resizebox{\columnwidth}{!}{
  \begin{tabular}{lccccccc}
    \toprule
    \textbf{Agent} & \textbf{Tools} & \textbf{Memory} & \textbf{Multi-Agent}  & \textbf{Customisable Control-Flow} & \textbf{SFT} & \textbf{RLFT} \\
    \midrule
    Transformers Agents\citep{huggingface_transformers_agent_2023} & 
    \textcolor{OliveGreen}{\large \checkmark} & \textcolor{BrickRed}{\large \texttimes} & \textcolor{BrickRed}{ \large \texttimes} &  \textcolor{BrickRed}{\large\texttimes}& \textcolor{BrickRed}{\large \texttimes} & \textcolor{BrickRed}{\large \texttimes} \\
    LangChain\citep{LangChain} & \textcolor{OliveGreen}{\large \checkmark} & \textcolor{OliveGreen}{\large\checkmark} & \textcolor{BrickRed}{\large\texttimes} &  \textcolor{BrickRed}{\large\texttimes}& \textcolor{BrickRed}{\large\texttimes} & \textcolor{BrickRed}{\large\texttimes} \\
    AutoGPT\citep{gravitas2023auto} & \textcolor{OliveGreen}{\large\checkmark} & \textcolor{OliveGreen}{\large\checkmark} & \textcolor{BrickRed}{\large\texttimes} &  \textcolor{BrickRed}{\large\texttimes}& \textcolor{BrickRed}{\large\texttimes} & \textcolor{BrickRed}{\large\texttimes} \\
    OpenAgents\citep{xie2023openagents} & \textcolor{OliveGreen}{\large\checkmark} & \textcolor{OliveGreen}{\large\checkmark} & \textcolor{BrickRed}{\large\texttimes} &  \textcolor{BrickRed}{\large\texttimes}& \textcolor{BrickRed}{\large\texttimes} & \textcolor{BrickRed}{\large\texttimes} \\
    XAgent\citep{xagent2023} & \textcolor{OliveGreen}{\large\checkmark} & \textcolor{OliveGreen}{\large\checkmark} & \textcolor{OliveGreen}{\large\checkmark} &  \textcolor{BrickRed}{\large\texttimes}& \textcolor{BrickRed}{\large\texttimes} & \textcolor{BrickRed}{\large\texttimes} \\
    MetaGPT\citep{hong2023metagpt} & \textcolor{OliveGreen}{\checkmark} & \textcolor{OliveGreen}{\large\checkmark} & \textcolor{OliveGreen}{\large\checkmark} &  \textcolor{BrickRed}{\large\texttimes}& \textcolor{BrickRed}{\large\texttimes} & \textcolor{BrickRed}{\large\texttimes} \\
    Camel\citep{li2023camel} & \textcolor{OliveGreen}{\large\checkmark} & \textcolor{OliveGreen}{\large\checkmark} & \textcolor{OliveGreen}{\large\checkmark} &  \textcolor{BrickRed}{\large\texttimes}& \textcolor{BrickRed}{\large\texttimes} & \textcolor{BrickRed}{\large\texttimes} \\
    AgentVerse\citep{chen2023agentverse} & \textcolor{OliveGreen}{\checkmark} & \textcolor{OliveGreen}{\large\checkmark} & \textcolor{OliveGreen}{\large\checkmark} &  \textcolor{BrickRed}{\large\texttimes}& \textcolor{BrickRed}{\large\texttimes} & \textcolor{BrickRed}{\large\texttimes} \\

    AGENTS\citep{zhou2023agents} & \textcolor{OliveGreen}{\large\checkmark} & \textcolor{OliveGreen}{\checkmark} & \textcolor{OliveGreen}{\large\checkmark} &  \textcolor{OliveGreen}{\large\checkmark}& \textcolor{BrickRed}{\large\texttimes} & \textcolor{BrickRed}{\large\texttimes} \\

    AutoGen\citep{wu2023autogen} & \textcolor{OliveGreen}{\large\checkmark} & \textcolor{OliveGreen}{\large\checkmark} & \textcolor{OliveGreen}{\large\checkmark} &  \textcolor{OliveGreen}{\large\checkmark}& \textcolor{BrickRed}{\large\texttimes} & \textcolor{BrickRed}{\large\texttimes} \\
    \midrule
    \textbf{Pangu-Agent} & \textcolor{OliveGreen}{\large\checkmark} & \textcolor{OliveGreen}{\large\checkmark} & \textcolor{OliveGreen}{\large\checkmark} &  \textcolor{OliveGreen}{\large\checkmark} & \textcolor{OliveGreen}{\large\checkmark} & \textcolor{OliveGreen}{\large\checkmark}\\
    \bottomrule
  \end{tabular}
  }
  \caption{Comparison of our proposed agent versus recent language-based agents. The overview of related methods can be found in \cref{sec:related}.}\label{tab:agent_summary} 
\end{table}
In short, we summarise our contributions as:
\begin{enumerate}
    \item We demonstrate the importance of structured reasoning in an agent framework that is general and versatile enough to effectively cover a diverse spectrum of existing agent framework in the literature~\citep[e.g.\ ][]{wang2022self,yao2022react,lin2023swiftsage,zhou2023leasttomost}. 
    As a meta-agent framework, it can be adjusted or fine-tuned utilising the sequence of intrinsic function calls or delegating decision-making to the underlying LLM. 
    As we show in \cref{appdx:nesting_examples}, users of the framework can easily expand the capabilities of the agent and combine or reuse many of the methods that have already been implemented.
    \item We conduct a comprehensive evaluation of structured reasoning by evaluating first-order (e.g.\ direct answering, chain-of-thought prompting~\citep{wei2022chain}) and composite methods (e.g. ReAct~\citep{yao2022react}, Self-Consistency~\citep{wang2022self}, SwiftSage~\citep{lin2023swiftsage}), on seven LLMs and six distinct domains. This evaluation can be used to inform researchers on how to initialise their agents and how to collect data for the fine-tuning steps. 
    \item  Our study demonstrates the impact of Supervised Fine-Tuning (SFT) and RL Fine-Tuning (RLFT) of the framework. Through structured reasoning, we successfully implement a rejection-sampling-based SFT pipeline, notably improving an LLM's performance in the ALFWorld domain~\citep{ALFWorld}, with success rates in held-out tasks increasing from 27\% to 82\%. Although the benefits of SFT plateau, we show that further enhancements are achievable through RL, increasing success rates to 88\% and even from 28\% to 91\% in the BabyAI task~\citep{chevalierboisvert2019babyai}. Moreover, our cross-domain experiments reveal the ability of a single LLM, trained via the RL pipeline, to \emph{simultaneously} achieve high performance in both the ALFWorld (82\%) and BabyAI (58.7\% average in 18 tasks) domains. These findings highlight the potential of structured reasoning to advance LLM-based agent training significantly.

\end{enumerate}

\section{The Pangu-Agent Formulation}\label{sec:formulation}

In this section, we continue our development of incorporating intrinsic functions into the RL pipeline to allow the definition of flexible reasoning processes. We emphasise the need for those to be defined, learned, and used separately from the extrinsic functions, whereby users can re-define any arbitrarily nesting deemed helpful for their tasks. We can rewrite the Pangu-Agent's optimisation problem from \cref{Eq:NewFormulation} in a more detailed form as: 
\begin{equation}
\label{Eq:NewFormulation2}
    \max_{\pi(\cdot)} \E_{\pi(\cdot)}\left[\sum_{t\geq 0}\gamma^t r_t\right] \rightarrow \overbrace{\max_{\pi(\cdot),\vec{\bm{\mu}}(\cdot)} \E_{\pi(\cdot),\vec{\bm{\mu}}(\cdot)}\left[\sum_{t\geq 0}\gamma^t r_t\right]}^{\text{Pangu Opt.}}, \ \ \text{with $\Vec{\bm{\mu}}(\cdot)$ being a set of intrinsic functions,}
\end{equation}
and where $r_t$ is the reward at the time step $t$ that depends on the environmental observation $\bm{o}_t$ and action $\bm{a}_t$. Furthermore, $\gamma\in[0,1)$ is a discount factor that specifies the degree to which rewards are discounted over time. The extrinsic functions still serve as actuators interacting with the outside world, whereas those additionally layered intrinsic functions are designed to encapsulate any internal reasoning process deemed beneficial by the system's architect.

Recent advancements from LLMs have demonstrated impressive successes across a broad spectrum of general reasoning and decision-making tasks to induce comprehensive prior knowledge. While fine-tunable, those language priors allow us to go beyond narrow task specifications and heavy engineering demands, and they also enable new paradigms and architectures, including, for example, memory management and tool usage, that can further aid in solving complex and challenging domains. Central to this design is the intrinsic function's role as a versatile encoder, whereby it can, for instance, facilitate the CoT mechanism \citep{wei2022chain,tutunov2023large}, utilising LLMs to generate and process internal thoughts, and it can serve as a dynamic memory or retrieval system, adept at accessing and utilising previously gathered information. In principle, we can introduce such intrinsic functions during any step of the RL pipeline. However, in this first version of our Pangu-Agent, we define our initial policies as (\emph{extrinsic}) functions of pre-trained LLMs, which can be later trained further since those action-selection rules are a central core component of the agents' interaction with environments. Faced with a task $\ell$ and equipped with a memory $\mem^{\ell}_{t}$, our agent decides to choose an action $\bm{a}_t$ as follows: 
\begin{equation}
\label{Eq:Policy}
\pi_{\text{Pangu-Agent}} \left(\bm{a}^{\ell}_{t}|\bm{o}^{\ell}_{1:t}, \mem^{\ell}_t\right) \equiv \mathcal{F}^{\ell}\left(\cdot \sim \llm\left( \phi^{\ell}\left(\bm{o}^{\ell}_{1:t},\mem^{\ell}_t\right)\right)\right).
\end{equation}
The above equation shows that Pangu-Agent chooses an action in a three-step nested process. First, it extracts relevant prompts for task $\ell$ using the observations so far and the current memory, i.e., the first step in the nesting $\phi^{\ell}(\bm{o}^{\ell}_{1:t}, \mem_{t}^{\ell})$ with $\phi^{\ell}(\cdot)$ denoting a task-relevant prompt generator. 
The generated prompt is passed to an LLM from which a response is sampled, i.e. $\cdot \sim \llm(\phi^{\ell}(\bm{o}^{\ell}_{1:t}, \mem_{t}^{\ell}))$.
Finally, the third step parses this response to enable compatible action execution in the environment via a task-specific action parsing functions $\mathcal{F}^{\ell}(\cdot)$, leading us to the Pangu-Agent's policy  $\pi_{\text{Pangu-Agent}}(\bm{a}^{\ell}_t|\bm{o}^{\ell}_{1:t},\mem_t^{\ell})$  described in \cref{Eq:Policy}.

Any entity that can learn and adapt must manage and form memories depending on past experiences and new environmental observations. Such memory components allow agents to draw upon past events to frame their understanding within the present context and can even aid in predicting and making sense of potential future events. To offer such flexibility, we do not restrict our agent to a specific form of memory but provide it with the ability to transform the internal state of the memory using a set of \emph{intrinsic functions}. Those (parameterisable) functions allow specific operations, e.g., using tools, planning, reflecting on past experience, communicating with other agents, etc. - see \cref{sec:arch} for a detailed exposition of such functions. Having decided on a nesting of those intrinsic functions, the agent's memory evolves to the next time step: 
\begin{equation*}
    \mem_t^{\ell} = \mu_{k}^{\ell}\left(\bm{o}_{1:t}^{\ell},\bm{a}_{1:t-1}^{\ell},\mu_{k-1}^{\ell}\left(\cdots \left(\mu_{1}^{\ell}\left(\bm{o}^{\ell}_{1:t},\bm{a}_{1:t-1}^{\ell},\mem^{\ell}_{t-1}\right)\right)\right)\right), \end{equation*}
with each $\mu^{\ell}_{k}$ being an intrinsic function. Therefore, instead of optimising the standard RL objective, Pangu-Agent attempts to find policies of the form presented in \cref{Eq:Policy} to maximise total discounted returns: 
\begin{align*}
    \max_{\pi_{\text{Pangu-Agent}}(), \bm{\mu}()} &\mathbb{E}_{\pi_{\text{Pangu-Agent}}(\cdot)}\left[\sum_{t\geq 0}\gamma^{t} r^{\ell}_t\right] \\
     \text{s.t.} \ \ &\pi_{\text{Pangu-Agent}} \left(\bm{a}^{\ell}_{t}|\bm{o}^{\ell}_{1:t}, \mem^{\ell}_t\right) = \mathcal{F}^{\ell}\left(\cdot \sim \llm\left( \phi^{\ell}\left(\bm{o}^{\ell}_{1:t},\mem^{\ell}_t\right)\right)\right) \ \forall t \\
    & \mem_t^{\ell} = \mu_{k}^{\ell}\left(\bm{o}_{1:t}^{\ell},\bm{a}_{1:t-1}^{\ell},\mu_{k-1}^{\ell}\left(\cdots \left(\mu_{1}^{\ell}\left(\bm{o}^{\ell}_{1:t},\bm{a}_{1:t-1}^{\ell},\mem^{\ell}_{t-1}\right)\right)\right)\right) \ \forall t,
\end{align*}
with $\Vec{\bm{\mu}}(\cdot)$ being the set of intrinsic functions that can be optimised in addition to the policy to maximise expected returns. This way, Pangu-Agent goes beyond other frameworks to support tunable and flexible learners at every step of the pipeline -- during policy training (or RL fine-tuning of LLMs) and the nested execution of intrinsic functions $\Vec{\bm{\mu}}(\cdot)$. The proposed framework, which includes structure constraints and memory manipulation (also known as test-time computation), overcomes the autoregressive computational limitations of LLMs \citep{lin2020limitations} and enables AI agents to achieve Turing completeness \citep{wei2022statistically}. 

\section{The Pangu-Agent Framework} 
 \label{sec:arch}
 
\subsection{Architecture of Pangu-Agent}

In this section, we investigate in depth the specifics of the architecture of our agent framework. As discussed in \cref{sec:formulation}, we first define two families of functions: \emph{intrinsic} and \emph{extrinsic}.

\textbf{Intrinsic Functions -- Operating on the Agent's Memory:}
Intrinsic functions are a family of functions that operate on the agent's memory state. These functions, denoted as \(\mu_k\), take as input the observation history \(\bm{o}_{1:t}\) and action history \(\bm{a}_{1:t-1}\) provided by the environment and the current memory state \(\mem_{t-1}\) of the agent from the previous time step. They then output a new memory state that incorporates the transformations applied by the intrinsic function.
The intrinsic functions are crucial in shaping the agent's internal state and can significantly impact its decision-making process. By leveraging these functions, the agent can adapt its memory state based on the observation history and previous knowledge, making more informed and contextually appropriate decisions.
Additionally, the intrinsic functions allow the incorporation of existing knowledge (through prompt engineering) and human-like structural thinking pipelines into the design of the AI agent.
Examples of such functions include \emph{thinking}, \emph{planning} and \emph{reflecting} on experience.
When asking the agent to think, it looks at the problem and produces a high-level thought about the current situation. 
However, creating agents does not end with the methods and techniques only performed within it. 
Agents are built to allow interaction with other agents and external tools. 
Therefore, two other important intrinsic functions are the \emph{communication} and \emph{tool-use}. 
First, by enabling pair-wise communication between agents, Pangu-Agent can simulate realistic interactions and close the sim-to-real gap. Second, tool use is a generic intrinsic function that allows the utilisation of a wide range of tools, such as code interpreters and web searching.

\textbf{Extrinsic Functions -- Interacting with the Environment:}
Extrinsic functions serve the purpose of eliciting an environment interaction from the language model. Unlike intrinsic functions that operate on the agent's memory state, extrinsic functions directly interact with the environment by generating actions to be executed.
These functions take as input the observation and action history \(\bm{o}_{1:t}, \bm{a}_{1:t-1}\) provided by the environment and the current memory state \(\mem^{\ell}_t\) of the agent. They utilise the transformed memory state obtained from the intrinsic functions and other contextual information to make informed decisions about the action that will maximise the agent's objective.

\textbf{Composite Methods -- SwiftSage~\citep{lin2023swiftsage}, ReAct~\citep{yao2022react}, Least-to-Most~\citep{zhou2023leasttomost}, and more:} The flexibility of our formulation means that many composite methods can be created hierarchically.
For instance, we give examples of the ReAct, Least-to-Most, and SwiftSage frameworks.
The ability to create such complex algorithms directly results from the modular formulation of Pangu-Agent. It does not mean that it is the only framework capable of these functions, but it is generic enough 
to support them. Additionally, it should be noted that the implementations of these composite methods that we provide in the Pangu-Agent codebase are not always faithful reproductions of the original algorithms, as these require specific task details. To allow for universal usage of these methods, we have adapted these methods to the Pangu-Agent framework by removing any task-specific details.
In \cref{sec:evaluation}, we present results for these methods in different tasks. We also show how simple it is to use and create these composite methods within the Pangu-Agent framework in \cref{appdx:nesting_examples}.

\textbf{Search-Enhanced Planning -- BFS, DFS, and MCTS}: Inspired by recent search-augmented LLMs \citep{yao2023tree, hao2023reasoning, feng2023alphazero}, Pangu-Agent framework integrates three tree-search algorithms -- Breadth-first/depth-first search (BFS/DFS) and Monte-Carlo Tree Search (MCTS), to increase the planning ability for better LLM's generation and decision-making ability. Specifically, Our framework leverages LLM as policy, model, and value functions. By interacting with this LLM-based simulated environment, we can construct a rollout tree which will be further pruned for better action/generation using tree-search algorithms. We conduct initial experiments on GSM8K and Game24 leveraging our framework shown in \cref{apx:planning} and we refer the readers to \citep{feng2023alphazero} for more detailed evaluation results.

\textbf{Task Interaction and Multi-Agent Systems:} Pangu-Agent is compatible with a range of tasks, for example, ALFWorld \citep{ALFWorld}, GSM8K \citep{cobbe2021training}, HotpotQA \citep{yang-etal-2018-hotpotqa}, WebShop \citep{yao2022webshop}, etc. 
The interface, influenced by OpenAI's Gym, is an open design that facilitates task interactions with any system where an agent performs actions and receives observations and rewards. Support for multi-agent environments is even built into the foundation of our framework, ensuring agents can work together or independently within the same task. 
In settings with multiple agents, we denote them with the subscript \(i\).
The interaction between one or more agents and an environment in our setting can be captured by the Partially Observable Stochastic Game~\citep{hansen2004dynamic} framework, which collapses to a Partially Observable Markov Decision Process in case there is only a single agent in the task.

\textbf{Prompting System:} The framework incorporates a template system to generate inputs for the LLMs.
Utilising templates enhances the flexibility of prompt crafting, making various prompting mechanisms possible. The system also promotes extensibility, with new prompting mechanisms or environments easily incorporated into the existing framework. Most intrinsic and extrinsic functions contain their task-specific prompting strategy, which uses the memory to create the prompt that will be inputted to the LLM.

\subsection{Adaptation of LLM Priors \& Fine-Tuning}\label{Sec:finetuning}
In introducing the architecture above, we aim for generalist agents that adapt based on expert data and environmental feedback using supervised fine-tuning and RL. We believe the interplay between structured reasoning from intrinsic functions and extrinsic policies promises general RL solvers that combine well with large model policies like those induced by LLMs. With our framework, its function design, and prompting strategies, we can now collect valuable rewarding trajectories from (open-sourced) pre-trained LLM priors to kickstart the training and fine-tuning process effectively. 

We differentiate between two types of fine-tuning methods: \textit{i}) those that do not interact with an environment, which we refer to as supervised fine-tuning algorithms, and \textit{ii)} those that learn by agent-environment interactions to maximise expected returns that we dub under RL-based fine-tuning techniques.    
Fine-tuning requires practical algorithms that update the weights of the LLM regardless of the type used. It is customary to define loss functions - e.g., predictive likelihood - and to rely on gradient descent when searching for update directions. However, computing those gradients for each weight of LLMs is often infeasible without accessing GPUs with vast memory capabilities. Hence, we support full-rank and low-rank adaptation algorithms to democratise and enable our agent's broader usage. This way, users can perform complete weight updates if their computing capabilities allow or rely otherwise on low-rank adaptation (LoRA) \citep{hu2021lora}. 
\paragraph{Fine-Tuning Type I:} To support the first type of fine-tuning (i.e., supervised), we allow the collection of successful trajectories from an environment via the different prompting strategies introduced before. Of course, we do not restrict our pipeline to data collection but also allow for the standard setup of introducing external expert datasets in case they are available. Given such a data set $\mathcal{D}$, we rely on causal language modelling losses - a special case of likelihood optimisation - and perform full- or low-rank updates to minimise:   
\begin{equation} \label{eq:4}
   \mathcal{L}(\mathcal{D}, \bm{\theta}_{\text{LLM}}) = -\frac{1}{N} \sum_{n=1}^N \text{log\,} p(x_n | x_{i \le n-1 }, \bm{\theta}_{\text{LLM}}),  
\end{equation}   
where $\bm{\theta}_{\text{LLM}}$ are the LLM's weights to be updated.
\Cref{eq:4} maximises the likelihood of the observed token $\bm{x}_{n}$ given all the previous ones. Since we are in an agent framework, our loss is only defined by the tokens generated by the agent and not the ones provided by the environment (or by the system prompts). Updates of $\bm{\theta}_{\text{LLM}}$ can be executed either fully if using full-rank optimisation or by parameterising them via two low-rank matrices in the LoRA scenario. Here, $\bm{\theta}_{LLM} = \bm{\theta}_{LLM}^{0}+\bm{B}\bm{A}$, where $\bm{\theta}_{LLM}^{0} \in \mathbb{R}^{d\times k}$ are the weights of the pre-trained LLM and $\bm{B} \in \mathbb{R}^{d \times r}$ and $\bm{A} \in \mathbb{R}^{r \times k}$ are tunable parameters of rank $r << \min\{d,k\}$. Here, we follow LoRA \citep{hu2021lora} in keeping $\bm{\theta}_{\text{LLM}}^{0}$ fixed and performing gradient updates only for $\bm{B}$ and $\bm{A}$.

\paragraph{Fine-Tuning Type II: } Regarding RL fine-tuners, we allow agents to interact with environments, as shown in \cref{Fig:RL}. Given a system prompt and access to an environment, the agent applies actions after which the environment transitions to a new state. This process repeats until the end of an episode, at which stage our learner receives a (sparse) reward signal. Given those rewards, we execute the PPO algorithm~\citep{PPO} to improve our agent's policy to maximise expected returns. 

Of course, this setup is standard for deep RL. However, we now elaborate on two crucial differences to deep RL. First, our agent uses intrinsic functions before deciding on an action via the policy. In our experiments, those intrinsic functions were relatively simple, whereby they ordered and arranged data from memory to generate suitable trajectories for extrinsic functions and shrank first observations and actions when exceeding context length constraints. Second, the vital difference with standard RL pipelines is that our agent utilises LLM policies. While such policies offer prior flexibility, they arrive with new generation speed and credit assignment problems if actions can be any text prompt.    

\paragraph{On Accelerating Generation:} A well-known bottleneck for successful RL with LLM policies is the speed by which LLMs generate actions. In an RL setting, slow-generation speeds coupled with the high-sample complexities typical for model-free algorithms lead to impractical learners, especially when only having access to modest computing hardware. We rely on three methods to accelerate generation speed: \textit{i)} continuous batching, \textit{ii)} optimised kernels and \textit{iii)} caching with Paged Attention \citep{kwon2023efficient}.
Since we collect trajectories in parallel with multiple workers interacting, each with its environment, continuous batching allows us to batch queries to the LLM and to possibly receive an answer before the whole batch computations are over.
Optimised attention kernels are written in C++/CUDA to enable faster memory-efficient inference than vanilla implementations.
We notably relied on the xFormers library \citep{xFormers2022}, which also features fused operations for the softmax, linear layer, dropout and layer normalisation operations.
Those fused operations notably reduce the amount of reading and writing back to the DRAM compared to vanilla PyTorch implementation.
Finally, we relied on Paged Attention \citep{kwon2023efficient} to cache the key and value in attention operations. 
Since the generation is autoregressive, it is unnecessary to recompute all the previous tokens' keys and values if they were computed before.
With those additions, we observed that we could reduce the generation time by a factor of 3 compared to vanilla HuggingFace for the Llama 2-7B parameters model. 

\paragraph{On Credit Assignment:} To work with environments where actions can be any textual prompt (e.g. in code generation) and in environments with a restricted set of actions, we would need to define the policy loss and the associated likelihood over every token composing an action. However, when defining critics in RL, credit must be assigned to an action, not a token. For instance, if the sequence of the following three tokens $[\left\langle \texttt{Move}\right\rangle, \left\langle \right\rangle, \left\langle \texttt{Right}\right\rangle]$ constitutes one action (i.e. that which moves the agent to the right), we would need to assign the same value for each of those three tokens. To support this form of credit assignment, we introduce a new version of generalised advantage estimation in PPO, where we define:
\begin{equation*}
    A_t = \sum_{n=t}^{\infty} \mathbbm{1}_{n \in \bar{\mathcal{A}}} (\lambda \gamma)^{\sigma(n)-\sigma(t)} \left(r_{\sigma(n)+1} + \gamma V(s_{\sigma(n)+1}) - V(s_{\sigma(n)}) \right),
\end{equation*}
where $\lambda$ is a hyper-parameter that balances the bias-variance tradeoff,
$\gamma$ is a discount factor that weighs the future returns,
$\bar{\mathcal{A}}$ is the set composed of the index of the last token of every action, and
$\sigma : \mathbb{N} \rightarrow \mathbb{N}$ is a mapping from the token index to the action index. 
Using $\bar{\mathcal{A}}$, we can keep track of the tokens corresponding to one action to perform the correct credit assignment. Finally, $V(s)$ is the current value estimate of a state $s$. We add a second head to the LLM from the last embedding layer to allow the critic to accurately approximate $V(s)$. We followed PPO \citep{PPO} to train this critic and minimise the squared temporal difference loss.

\section{Evaluating Pangu-Agent} \label{sec:evaluation}
\begin{figure}[t]
    \centering
    \includegraphics[trim={0em 9em 0em 7em },clip,width=.8\textwidth]{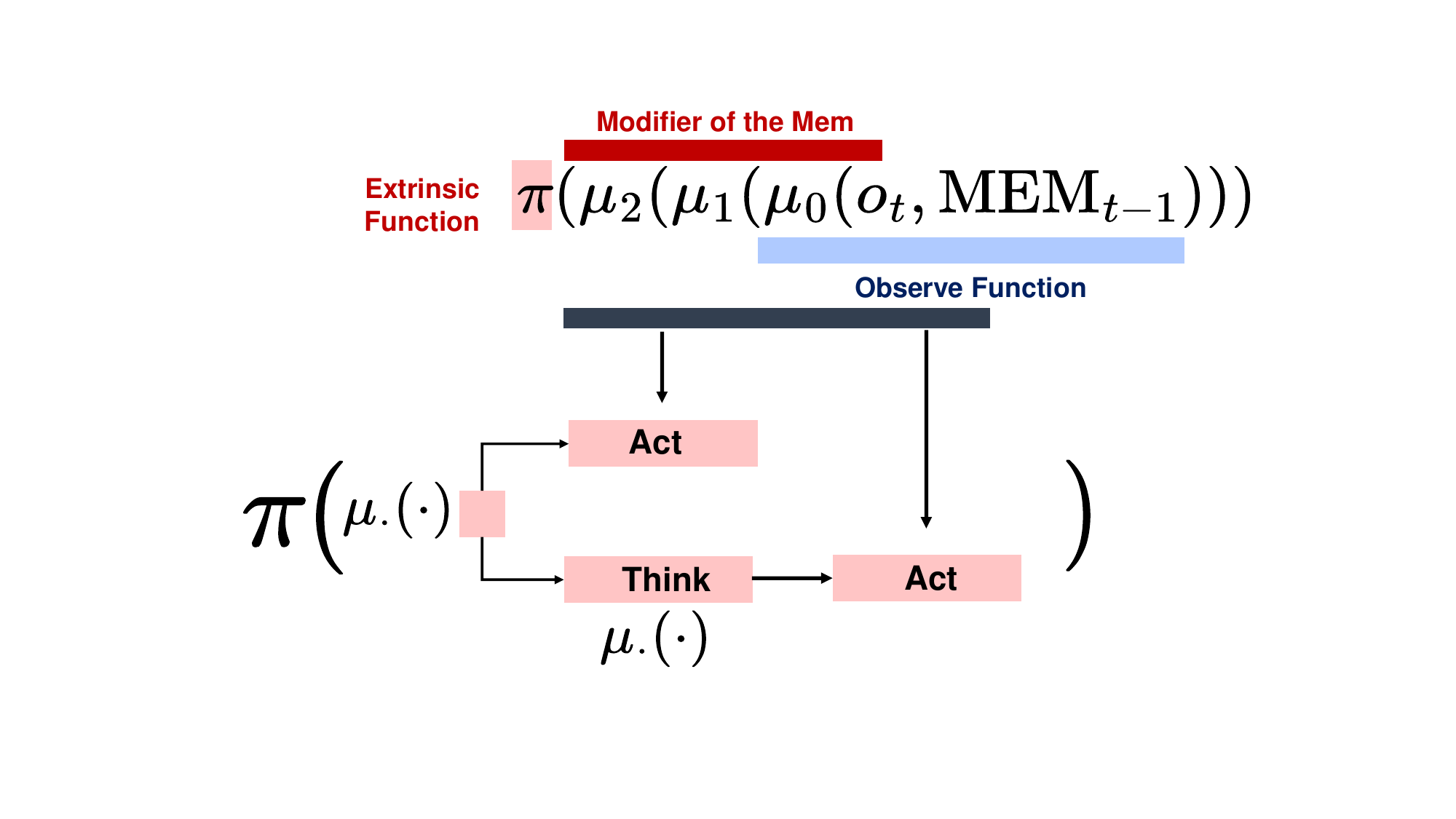}
    \caption{Visualisation of one example of structured reasoning using nesting of intrinsic and extrinsic functions. The agent initially updates its internal memory, using $\mu_0$, by perceiving its observation. Then the intrinsic function $\mu_1$ selects between Think-and-Act or just Act. The last intrinsic function $\mu_2$ either generates a thought if $\mu_1$ previously selected Think-and-Act otherwise it is null. Finally, the extrinsic function $\pi$ selects the action that the agent will perform in the environment.}
    \label{fig:structured_reasoning}
\end{figure}

In this section, we conduct an extensive evaluation of various methods supported by  Pangu-Agent, including both structured reasoning methods and fine-tuning. First, we evaluate the structured reasoning ability of Pangu-Agent (see \Cref{fig:structured_reasoning}, by considering both first-order nesting and composite methods, and afterwards we evaluate fine-tuning of Pangu-Agent in three distinct settings using supervised learning and RL. Our results indicate that composite methods tend to outperform first-order nesting methods with respect to the achieved returns of the agent. Finally, we show that SFT and RLFT allow the agent to specialise and further increase its achieved returns in the ALFWorld and BabyAI tasks.
Throughout this evaluation, several LLMs are used, such as GPT \citep{openai2023gpt4}, Llama 2 \citep{llama2}, OpenChat \citep{openchat}, Vicuna \citep{vicuna}, and Mistral \citep{mistral7b}.

\subsection{Evaluating Structured Reasoning}\label{eval:prompt_engineering}
\label{sec:exp_pm}

The built-in support for intrinsic functions allows us to evaluate how different design choices in the reasoning structure affect the performance of an AI agent. 
First, in \Cref{tab:results:first-order}, we evaluate first-order nestings, i.e.\ setups in which an agent's memory is modified only by observing the environment and the actions it performs on it. In the literature, these would be referred to simply as different \emph{prompting methods} e.g.: Direct prompting, prompting with Few-Shot (FS) examples, Few-Shot Chain of Thought (FS-CoT)~\citep{wei2022chain}, Zero-Shot CoT (ZS-CoT)~\citep{NEURIPS2022_8bb0d291} -- with a thorough description of these methods presented in \cref{appdx:intrinsic-extrinsic}.  It should be noted that due to the non-deterministic nature of LLM's text generation, the achieved returns can vary significantly between different runs. To account for these variations, we run each combination of task-method-LLM three times and report the mean standard deviation. 

\begin{table}[t]
\resizebox{\columnwidth}{!}{\begin{tabular}{@{}clccccccc@{}}
\toprule
\multirow{2}{*}{\textbf{LLM}}      & \multicolumn{1}{c}{\multirow{2}{*}{\textbf{Method}}} & \multirow{2}{*}{\textbf{Overall}} & \multicolumn{6}{c}{\textbf{Task}}                                                                                                 \\ \cmidrule(l){4-9} 
                                   & \multicolumn{1}{c}{}                                 &                                   & \textbf{ALFWorld}   & \textbf{GSM8K}      & \textbf{HotpotQA}   & \textbf{WebShop}    & \textbf{HumanEval}  & \textbf{BabyAI}     \\ \hline \\[-2ex]
\multirow{4}{*}{\rotatebox[origin=c]{90}{\scriptsize{\textbf{GPT-3.5}}}}
& \textbf{Direct           } & \textbf{35.9}             & 4.7 \small{$ \pm \ 2.5$}  & 69.2 \small{$ \pm \ 0.8$} & 32.8 \small{$ \pm \ 0.7$} & 22.5 \small{$ \pm \ 3.5$} & 58.2 \small{$ \pm \ 2.1$} & 28.3 \small{$ \pm \ 4.7$} \\
& \textbf{ZS-CoT           } & \textbf{29.3}             & 24.7 \small{$ \pm \ 1.9$} & 65.8 \small{$ \pm \ 0.9$} & 25.8 \small{$ \pm \ 0.1$} & 6.9 \small{$ \pm \ 1.3$}  & 17.8 \small{$ \pm \ 0.8$} & 34.6 \small{$ \pm \ 0.9$} \\
& \textbf{FS               } & \textbf{38.3}             & 12.7 \small{$ \pm \ 3.8$} & 35.0 \small{$ \pm \ 11.5$} & 45.3 \small{$ \pm \ 0.7$} & 34.4 \small{$ \pm \ 1.2$} & 66.5 \small{$ \pm \ 3.5$} & 36.2 \small{$ \pm \ 2.6$} \\
& \textbf{FS-CoT           } & \textbf{49.9}             & 40.3 \small{$ \pm \ 3.1$} & 66.4 \small{$ \pm \ 0.1$} & 40.5 \small{$ \pm \ 0.3$} & 38.9 \small{$ \pm \ 1.7$} & 63.1 \small{$ \pm \ 3.7$} & 50.2 \small{$ \pm \ 4.3$} \\
\midrule

\multirow{4}{*}{\rotatebox[origin=c]{90}{\scriptsize{\textbf{Llama 2-70B}}}}
& \textbf{Direct           } & \textbf{17.9}             & 5.3 \small{$ \pm \ 0.9$}  & 28.8 \small{$ \pm \ 0.7$} & 27.6 \small{$ \pm \ 0.4$} & 0.0 \small{$ \pm \ 0.0$}  & 23.2 \small{$ \pm \ 1.5$} & 22.2 \small{$ \pm \ 2.0$} \\
& \textbf{ZS-CoT           } & \textbf{21.0}             & 11.3 \small{$ \pm \ 0.9$} & 48.7 \small{$ \pm \ 1.0$} & 33.9 \small{$ \pm \ 0.5$} & 6.6 \small{$ \pm \ 2.2$}  & 11.8 \small{$ \pm \ 2.0$} & 13.7 \small{$ \pm \ 1.9$} \\
& \textbf{FS               } & \textbf{16.3}             & 5.3 \small{$ \pm \ 3.4$}  & 32.2 \small{$ \pm \ 0.7$} & 31.0 \small{$ \pm \ 0.4$} & 1.2 \small{$ \pm \ 0.8$}  & 15.7 \small{$ \pm \ 1.1$} & 12.3 \small{$ \pm \ 3.7$} \\
& \textbf{FS-CoT           } & \textbf{27.7}             & 18.0 \small{$ \pm \ 4.3$} & 53.0 \small{$ \pm \ 0.6$} & 49.0 \small{$ \pm \ 0.6$} & 5.7 \small{$ \pm \ 1.6$}  & 19.0 \small{$ \pm \ 0.3$} & 21.6 \small{$ \pm \ 3.1$} \\
\midrule 

\multirow{4}{*}{\rotatebox[origin=c]{90}{\scriptsize{\textbf{OpenChat-3.2}}}}
& \textbf{Direct           } & \textbf{11.5}             & 0.0 \small{$ \pm \ 0.0$}  & 7.6 \small{$ \pm \ 0.6$}  & 43.0 \small{$ \pm \ 0.4$} & 0.0 \small{$ \pm \ 0.0$}  & 8.5 \small{$ \pm \ 2.1$}  & 9.8 \small{$ \pm \ 4.2$}  \\
& \textbf{ZS-CoT           } & \textbf{14.8}             & 17.3 \small{$ \pm \ 6.6$} & 22.7 \small{$ \pm \ 0.5$} & 28.6 \small{$ \pm \ 0.2$} & 0.0 \small{$ \pm \ 0.0$}  & 2.5 \small{$ \pm \ 0.9$}  & 17.7 \small{$ \pm \ 4.6$} \\
& \textbf{FS               } & \textbf{14.0}             & 2.0 \small{$ \pm \ 1.6$}  & 2.2 \small{$ \pm \ 0.4$}  & 47.0 \small{$ \pm \ 0.3$} & 22.2 \small{$ \pm \ 3.0$} & 1.4 \small{$ \pm \ 0.6$}  & 9.0 \small{$ \pm \ 6.0$}  \\
& \textbf{FS-CoT           } & \textbf{21.4}             & 26.7 \small{$ \pm \ 4.1$} & 21.1 \small{$ \pm \ 0.9$} & 39.2 \small{$ \pm \ 0.3$} & 3.2 \small{$ \pm \ 2.5$}  & 1.2 \small{$ \pm \ 0.0$}  & 36.8 \small{$ \pm \ 11.5$} \\
\midrule

\multirow{4}{*}{\rotatebox[origin=c]{90}{\scriptsize{\textbf{Vicuna-13B}}}}
& \textbf{Direct           } & \textbf{10.0}             & 3.3 \small{$ \pm \ 2.5$}  & 10.8 \small{$ \pm \ 0.2$} & 24.9 \small{$ \pm \ 0.5$} & 0.0 \small{$ \pm \ 0.0$}  & 0.0 \small{$ \pm \ 0.0$}  & 21.0 \small{$ \pm \ 2.2$} \\
& \textbf{ZS-CoT           } & \textbf{13.3}             & 13.3 \small{$ \pm \ 1.9$} & 23.8 \small{$ \pm \ 0.3$} & 19.3 \small{$ \pm \ 0.5$} & 0.4 \small{$ \pm \ 0.3$}  & 0.2 \small{$ \pm \ 0.3$}  & 22.9 \small{$ \pm \ 2.8$} \\
& \textbf{FS               } & \textbf{14.7}             & 11.3 \small{$ \pm \ 2.5$} & 6.2 \small{$ \pm \ 0.6$}  & 30.5 \small{$ \pm \ 0.4$} & 12.2 \small{$ \pm \ 0.7$} & 1.2 \small{$ \pm \ 0.9$}  & 27.0 \small{$ \pm \ 2.3$} \\
& \textbf{FS-CoT           } & \textbf{17.8}             & 15.3 \small{$ \pm \ 0.9$} & 27.3 \small{$ \pm \ 0.9$} & 32.1 \small{$ \pm \ 0.9$} & 1.0 \small{$ \pm \ 0.8$}  & 2.1 \small{$ \pm \ 1.1$}  & 29.2 \small{$ \pm \ 2.9$} \\
\midrule 

\multirow{4}{*}{\rotatebox[origin=c]{90}{\scriptsize{\textbf{Llama 2-13B}}}}
& \textbf{Direct           } & \textbf{11.3}             & 0.0 \small{$ \pm \ 0.0$}  & 16.7 \small{$ \pm \ 0.6$} & 27.6 \small{$ \pm \ 0.4$} & 3.5 \small{$ \pm \ 1.2$}  & 11.6 \small{$ \pm \ 1.3$} & 8.7 \small{$ \pm \ 5.6$}  \\
& \textbf{ZS-CoT           } & \textbf{13.5}             & 0.7 \small{$ \pm \ 0.9$}  & 31.6 \small{$ \pm \ 0.6$} & 25.4 \small{$ \pm \ 0.4$} & 0.4 \small{$ \pm \ 0.6$}  & 9.7 \small{$ \pm \ 0.3$}  & 12.9 \small{$ \pm \ 3.7$} \\
& \textbf{FS               } & \textbf{12.1}             & 0.0 \small{$ \pm \ 0.0$}  & 7.9 \small{$ \pm \ 0.1$}  & 35.4 \small{$ \pm \ 0.7$} & 10.9 \small{$ \pm \ 1.5$} & 8.9 \small{$ \pm \ 1.1$}  & 9.8 \small{$ \pm \ 2.3$}  \\
& \textbf{FS-CoT           } & \textbf{16.4}             & 6.0 \small{$ \pm \ 1.6$}  & 33.5 \small{$ \pm \ 0.5$} & 23.9 \small{$ \pm \ 0.9$} & 11.2 \small{$ \pm \ 1.2$} & 2.1 \small{$ \pm \ 0.3$}  & 21.5 \small{$ \pm \ 4.2$} \\
\midrule 

\multirow{4}{*}{\rotatebox[origin=c]{90}{\scriptsize{\textbf{OpenChat-3.5}}}}
& \textbf{Direct           } & \textbf{30.9}             & 4.0 \small{$ \pm \ 0.0$}  & 37.6 \small{$ \pm \ 0.6$} & 47.4 \small{$ \pm \ 0.4$} & 35.4 \small{$ \pm \ 0.8$} & 32.9 \small{$ \pm \ 0.0$} & 28.1 \small{$ \pm \ 0.6$} \\
& \textbf{ZS-CoT           } & \textbf{34.9}             & 10.7 \small{$ \pm \ 5.7$} & 64.9 \small{$ \pm \ 0.4$} & 41.2 \small{$ \pm \ 1.0$} & 17.8 \small{$ \pm \ 1.4$} & 39.1 \small{$ \pm \ 4.3$} & 35.7 \small{$ \pm \ 1.0$} \\
& \textbf{FS               } & \textbf{31.2}             & 7.3 \small{$ \pm \ 2.5$}  & 24.7 \small{$ \pm \ 1.1$} & 57.2 \small{$ \pm \ 0.5$} & 43.5 \small{$ \pm \ 1.0$} & 33.3 \small{$ \pm \ 3.0$} & 20.9 \small{$ \pm \ 6.7$} \\
& \textbf{FS-CoT           } & \textbf{47.4}             & 27.3 \small{$ \pm \ 3.8$} & 70.1 \small{$ \pm \ 0.5$} & 64.3 \small{$ \pm \ 0.3$} & 35.5 \small{$ \pm \ 2.1$} & 28.0 \small{$ \pm \ 3.2$} & 58.9 \small{$ \pm \ 6.7$} \\
\midrule
                                   
\multirow{4}{*}{\rotatebox[origin=c]{90}{\scriptsize{\textbf{Mistral-7B}}}}
& \textbf{Direct           } & \textbf{11.2}             & 4.0 \small{$ \pm \ 1.6$}  & 14.1 \small{$ \pm \ 0.7$} & 30.3 \small{$ \pm \ 0.1$} & 0.0 \small{$ \pm \ 0.0$}  & 10.8 \small{$ \pm \ 0.6$} & 7.7 \small{$ \pm \ 3.2$}  \\
& \textbf{ZS-CoT           } & \textbf{13.2}             & 9.3 \small{$ \pm \ 3.8$}  & 33.9 \small{$ \pm \ 0.7$} & 18.3 \small{$ \pm \ 0.5$} & 0.0 \small{$ \pm \ 0.0$}  & 0.4 \small{$ \pm \ 0.3$}  & 17.4 \small{$ \pm \ 4.4$} \\
& \textbf{FS               } & \textbf{8.3}              & 6.0 \small{$ \pm \ 1.6$}  & 0.5 \small{$ \pm \ 0.0$}  & 32.9 \small{$ \pm \ 0.5$} & 0.7 \small{$ \pm \ 0.5$}  & 3.7 \small{$ \pm \ 0.5$}  & 6.0 \small{$ \pm \ 3.3$}  \\
& \textbf{FS-CoT           } & \textbf{17.5}             & 6.0 \small{$ \pm \ 2.5$}  & 32.7 \small{$ \pm \ 0.3$} & 28.0 \small{$ \pm \ 1.1$} & 0.9 \small{$ \pm \ 0.4$}  & 8.9 \small{$ \pm \ 2.4$}  & 28.6 \small{$ \pm \ 3.9$} \\

\bottomrule

\end{tabular}}
\caption{Average achieved returns and the standard deviation across three runs, for four first-order prompt engineering methods on six different tasks, using seven different LLMs.}
\label{tab:results:first-order}
\end{table}

But first-order nestings have limits as they may struggle to fully use the capabilities of an LLM. As motivated in the introduction (\cref{sec:introduction}), an agent needs to be able to process the output of the language model, revisit its answers, change its memory and even use tools. Composite methods, as we call them, are methods that may require multiple thinking steps until a final action is decided. In \Cref{tab:results:composite}, we present results for four composite methods: FS-CoT with Self-Consistency (FS-CoT-SC)~\citep{wang2022self}, FS-CoT with an optional distinct thinking step (e.g.\ React~\citep{yao2022react}), FS-CoT with a reflection step~\citep[e.g.\ ][]{shinn2023reflexion}, SwiftSage~\citep{lin2023swiftsage}, and Least-to-Most~\citep{zhou2023leasttomost} (also see \cref{appdx:composite}). All these methods make use of multiple intrinsic function steps at each environment time step. Refer to \cref{tab:methods_summary} for brief descriptions of all method acronyms.

\begin{table}[t]
\resizebox{\columnwidth}{!}{\begin{tabular}{@{}clccccccc@{}}
\toprule
\multirow{2}{*}{\textbf{LLM}}      & \multicolumn{1}{c}{\multirow{2}{*}{\textbf{Method}}} & \multirow{2}{*}{\textbf{Overall}} & \multicolumn{6}{c}{\textbf{Task}}                                                                                               \\ \cmidrule(l){4-9} 
                                   & \multicolumn{1}{c}{}                                 &                                   & \textbf{ALFWorld}   & \textbf{GSM8K}      & \textbf{HotpotQA}   & \textbf{WebShop}   & \textbf{HumanEval} & \textbf{BabyAI}     \\ \hline \\ [-2ex]
\multirow{4}{*}{\rotatebox[origin=c]{90}{\small{\textbf{GPT-3.5}}}}
& \textbf{FS-CoT-SC        } & \textbf{48.2}             & 34.4 \small{$ \pm \ 2.7$} & 74.1 \small{$ \pm \ 0.8$} & 43.5 \small{$ \pm \ 0.1$} & 39.1 \small{$ \pm \ 1.9$} & -                         & 50.0 \small{$ \pm \ 7.9$} \\
& \textbf{FS-CoT-React     } & \textbf{45.8}             & 39.5 \small{$ \pm \ 3.6$} & 66.9 \small{$ \pm \ 1.2$} & 38.1 \small{$ \pm \ 0.1$} & 28.5 \small{$ \pm \ 3.4$} & 61.3 \small{$ \pm \ 2.8$} & 40.8 \small{$ \pm \ 2.3$} \\
& \textbf{FS-CoT-Reflect   } & \textbf{34.0}             & 26.7 \small{$ \pm \ 1.9$} & 70.8 \small{$ \pm \ 0.8$} & 32.5 \small{$ \pm \ 0.9$} & 0.9 \small{$ \pm \ 0.4$}  & 54.5 \small{$ \pm \ 3.6$} & 19.0 \small{$ \pm \ 2.5$} \\
& \textbf{FS-Least-to-Most } & \textbf{39.8}             & -                         & 60.0 \small{$ \pm \ 0.3$} & 20.3 \small{$ \pm \ 0.4$} & -                         & 39.1 \small{$ \pm \ 1.3$} & -                         \\
\midrule

\multirow{5}{*}{\rotatebox[origin=c]{90}{\small{\textbf{Llama 2-70B}}}}
& \textbf{FS-CoT-SC        } & \textbf{29.7}             & 13.3 \small{$ \pm \ 2.5$} & 59.7 \small{$ \pm \ 0.8$} & 52.1 \small{$ \pm \ 0.3$} & 2.2 \small{$ \pm \ 0.3$}  & -                         & 21.0 \small{$ \pm \ 1.3$} \\
& \textbf{FS-CoT-React     } & \textbf{21.0}             & 3.3 \small{$ \pm \ 1.9$}  & 48.7 \small{$ \pm \ 1.1$} & 41.8 \small{$ \pm \ 0.8$} & 0.7 \small{$ \pm \ 1.0$}  & 12.6 \small{$ \pm \ 1.5$} & 19.0 \small{$ \pm \ 2.6$} \\
& \textbf{FS-CoT-Reflect   } & \textbf{24.6}             & 18.7 \small{$ \pm \ 3.4$} & 56.8 \small{$ \pm \ 1.1$} & 35.8 \small{$ \pm \ 0.8$} & 1.9 \small{$ \pm \ 1.2$}  & 19.0 \small{$ \pm \ 0.8$} & 15.5 \small{$ \pm \ 2.1$} \\
& \textbf{FS-CoT-Swift-Sage} & \textbf{23.5}             & 28.0 \small{$ \pm \ 5.7$} & -                         & -                         & 15.4 \small{$ \pm \ 0.9$} & -                         & 27.2 \small{$ \pm \ 2.1$} \\
& \textbf{FS-Least-to-Most } & \textbf{16.2}             & -                         & 31.2 \small{$ \pm \ 0.5$} & 15.0 \small{$ \pm \ 0.2$} & -                         & 2.5 \small{$ \pm \ 1.3$}  & -                         \\
\midrule

\multirow{5}{*}{\rotatebox[origin=c]{90}{\small{\textbf{OpenChat-3.2}}}}
& \textbf{FS-CoT-SC        } & \textbf{25.3}             & 21.3 \small{$ \pm \ 1.9$} & 21.2 \small{$ \pm \ 0.6$} & 42.7 \small{$ \pm \ 0.6$} & 1.3 \small{$ \pm \ 0.9$}  & -                         & 40.0 \small{$ \pm \ 3.6$} \\
& \textbf{FS-CoT-React     } & \textbf{11.3}             & 2.7 \small{$ \pm \ 0.9$}  & 6.6 \small{$ \pm \ 0.7$}  & 42.6 \small{$ \pm \ 1.0$} & 0.6 \small{$ \pm \ 0.5$}  & 3.3 \small{$ \pm \ 1.5$}  & 11.9 \small{$ \pm \ 2.4$} \\
& \textbf{FS-CoT-Reflect   } & \textbf{16.7}             & 20.7 \small{$ \pm \ 9.3$} & 26.2 \small{$ \pm \ 0.8$} & 25.2 \small{$ \pm \ 0.3$} & 0.0 \small{$ \pm \ 0.0$}  & 3.9 \small{$ \pm \ 0.8$}  & 24.3 \small{$ \pm \ 0.8$} \\
& \textbf{FS-CoT-Swift-Sage} & \textbf{21.3}             & 24.0 \small{$ \pm \ 6.9$} & -                         & -                         & 13.2 \small{$ \pm \ 2.2$} & -                         & 26.6 \small{$ \pm \ 3.1$} \\
& \textbf{FS-Least-to-Most } & \textbf{8.6}              & -                         & 14.0 \small{$ \pm \ 0.6$} & 11.3 \small{$ \pm \ 0.6$} & -                         & 0.4 \small{$ \pm \ 0.6$}  & -                         \\
\midrule

\multirow{5}{*}{\rotatebox[origin=c]{90}{\small{\textbf{Vicuna-13B}}}}
& \textbf{FS-CoT-SC        } & \textbf{25.6}             & 18.7 \small{$ \pm \ 2.5$} & 37.7 \small{$ \pm \ 0.5$} & 38.2 \small{$ \pm \ 0.2$} & 0.0 \small{$ \pm \ 0.0$}  & -                         & 33.4 \small{$ \pm \ 0.9$} \\
& \textbf{FS-CoT-React     } & \textbf{16.1}             & 17.3 \small{$ \pm \ 1.9$} & 19.0 \small{$ \pm \ 1.1$} & 26.0 \small{$ \pm \ 1.4$} & 0.0 \small{$ \pm \ 0.0$}  & 1.2 \small{$ \pm \ 0.9$}  & 33.0 \small{$ \pm \ 7.5$} \\
& \textbf{FS-CoT-Reflect   } & \textbf{16.9}             & 17.3 \small{$ \pm \ 3.4$} & 32.8 \small{$ \pm \ 1.0$} & 21.7 \small{$ \pm \ 0.4$} & 0.3 \small{$ \pm \ 0.4$}  & 4.8 \small{$ \pm \ 1.8$}  & 24.3 \small{$ \pm \ 4.2$} \\
& \textbf{FS-CoT-Swift-Sage} & \textbf{22.5}             & 27.3 \small{$ \pm \ 3.4$} & -                         & -                         & 13.9 \small{$ \pm \ 2.2$} & -                         & 26.2 \small{$ \pm \ 4.1$} \\
& \textbf{FS-Least-to-Most } & \textbf{10.0}             & -                         & 19.7 \small{$ \pm \ 0.4$} & 10.1 \small{$ \pm \ 0.3$} & -                         & 0.2 \small{$ \pm \ 0.3$}  & -                         \\
\midrule

\multirow{5}{*}{\rotatebox[origin=c]{90}{\small{\textbf{Llama 2-13B}}}}
& \textbf{FS-CoT-SC        } & \textbf{20.1}             & 3.3 \small{$ \pm \ 0.9$}  & 39.3 \small{$ \pm \ 0.6$} & 40.8 \small{$ \pm \ 0.7$} & 2.7 \small{$ \pm \ 1.6$}  & -                         & 14.5 \small{$ \pm \ 4.5$} \\
& \textbf{FS-CoT-React     } & \textbf{16.1}             & 0.0 \small{$ \pm \ 0.0$}  & 25.9 \small{$ \pm \ 0.5$} & 39.0 \small{$ \pm \ 0.5$} & 2.1 \small{$ \pm \ 0.9$}  & 3.7 \small{$ \pm \ 0.9$}  & 25.6 \small{$ \pm \ 7.5$} \\
& \textbf{FS-CoT-Reflect   } & \textbf{12.8}             & 10.7 \small{$ \pm \ 2.5$} & 32.4 \small{$ \pm \ 0.8$} & 11.0 \small{$ \pm \ 0.4$} & 0.3 \small{$ \pm \ 0.4$}  & 2.1 \small{$ \pm \ 0.6$}  & 20.6 \small{$ \pm \ 2.5$} \\
& \textbf{FS-CoT-Swift-Sage} & \textbf{18.3}             & 22.7 \small{$ \pm \ 0.9$} & -                         & -                         & 11.8 \small{$ \pm \ 4.9$} & -                         & 20.6 \small{$ \pm \ 2.5$} \\
& \textbf{FS-Least-to-Most } & \textbf{8.7}              & -                         & 12.2 \small{$ \pm \ 0.8$} & 13.1 \small{$ \pm \ 0.5$} & -                         & 0.8 \small{$ \pm \ 0.8$}  & -                         \\
\midrule

\multirow{5}{*}{\rotatebox[origin=c]{90}{\small{\textbf{OpenChat-3.5}}}}
& \textbf{FS-CoT-SC        } & \textbf{53.5}             & 28.0 \small{$ \pm \ 1.6$} & 80.8 \small{$ \pm \ 0.2$} & 70.4 \small{$ \pm \ 0.7$} & 42.9 \small{$ \pm \ 2.0$} & 32.9 \small{$ \pm \ 0.9$} & 66.1 \small{$ \pm \ 6.4$} \\
& \textbf{FS-CoT-React     } & \textbf{39.0}             & 24.7 \small{$ \pm \ 1.9$} & 62.1 \small{$ \pm \ 0.3$} & 48.2 \small{$ \pm \ 1.0$} & 33.6 \small{$ \pm \ 0.7$} & 29.2 \small{$ \pm \ 1.8$} & 36.2 \small{$ \pm \ 3.6$} \\
& \textbf{FS-CoT-Reflect   } & \textbf{39.4}             & 28.7 \small{$ \pm \ 5.7$} & 74.5 \small{$ \pm \ 0.2$} & 57.0 \small{$ \pm \ 0.1$} & 14.1 \small{$ \pm \ 3.2$} & 27.1 \small{$ \pm \ 3.6$} & 34.8 \small{$ \pm \ 1.2$} \\
& \textbf{FS-CoT-Swift-Sage} & \textbf{37.7}             & 37.3 \small{$ \pm \ 6.2$} & -                         & -                         & 31.5 \small{$ \pm \ 1.4$} & -                         & 44.3 \small{$ \pm \ 3.5$} \\
& \textbf{FS-Least-to-Most } & \textbf{36.0}             & -                         & 59.1 \small{$ \pm \ 0.6$} & 32.5 \small{$ \pm \ 0.1$} & -                         & 16.4 \small{$ \pm \ 0.6$} & -                         \\
\midrule

\multirow{5}{*}{\rotatebox[origin=c]{90}{\small{\textbf{Mistral-7B}}}}
& \textbf{FS-CoT-SC        } & \textbf{21.3}             & 6.0 \small{$ \pm \ 1.6$}  & 38.2 \small{$ \pm \ 0.7$} & 37.8 \small{$ \pm \ 0.4$} & 0.3 \small{$ \pm \ 0.4$}  & -                         & 24.4 \small{$ \pm \ 3.3$} \\
& \textbf{FS-CoT-React     } & \textbf{11.4}             & 5.3 \small{$ \pm \ 1.9$}  & 19.9 \small{$ \pm \ 1.1$} & 28.9 \small{$ \pm \ 0.9$} & 0.6 \small{$ \pm \ 0.8$}  & 5.8 \small{$ \pm \ 3.1$}  & 7.9 \small{$ \pm \ 2.0$}  \\
& \textbf{FS-CoT-Reflect   } & \textbf{15.0}             & 7.3 \small{$ \pm \ 0.9$}  & 34.4 \small{$ \pm \ 1.4$} & 19.6 \small{$ \pm \ 0.9$} & 0.0 \small{$ \pm \ 0.0$}  & 5.2 \small{$ \pm \ 0.3$}  & 23.3 \small{$ \pm \ 2.5$} \\
& \textbf{FS-CoT-Swift-Sage} & \textbf{19.4}             & 16.7 \small{$ \pm \ 5.0$} & -                         & -                         & 12.4 \small{$ \pm \ 3.0$} & -                         & 29.2 \small{$ \pm \ 2.8$} \\
& \textbf{FS-Least-to-Most } & \textbf{5.6}              & -                         & 5.5 \small{$ \pm \ 0.7$}  & 10.8 \small{$ \pm \ 0.3$} & -                         & 0.6 \small{$ \pm \ 0.5$}  & -                         \\
\bottomrule

\end{tabular}}
\caption{Average achieved returns and the standard deviation across three runs, for five composite reasoning methods on six different tasks, using seven different LLMs.}
\label{tab:results:composite}
\end{table}

We observe that methods that are similar in their structure but differ in their prompt content, such as Reflect and React, yield significantly different achieved returns for the agent. This demonstrates the importance of careful prompt engineering. It is also noteworthy that different methods work better for some LLMs than others, e.g. React in OpenChat-3.2 performs worse than FS on average, while React and FS in GPT-3.5 perform similarly in terms of average achieved returns.

Notably, the performance of FS in GSM8K is considerably worse than Direct across all LLMs. 
This does not come as a surprise since FS presents only the final answer to the LLM. 
Therefore, the LLM aims to answer the question without generating the intermediate steps. 
However, in Direct, the LLM generates the intermediate steps even without explicitly requested as this is how similar grade-school level questions are presented on the internet, which are likely contained within the training set of these LLMs. Similar conclusions can be drawn when comparing ZS-CoT with FS, where we observe that the achieved returns are increased even compared to Direct. This is especially true for smaller LLMs where we conjecture that when adding the "think step-by-step" quote into the prompt, the model is more likely to generate reasoning steps that will allow correctly solving the question at hand.

In the HumanEval task, we observe that the difference in the achieved returns between GPT-3.5 and the remaining models is significantly larger compared to other tasks. This can be attributed to the fact that HumanEval is a coding task, that requires well-structured responses, such as correct indentation, from the LLM. However, smaller and open-source LLMs are more prone to these structural errors which result in failing the task and receiving a return of 0.

Another factor that impedes the performance of LLMs is the limited context length. In tasks such as WebShop, which involves relatively large observations, the length of the prompt needs to be truncated to stay within the allowed context length. Consequently, the performance of LLMs in this task can be substantially affected, particularly in methods such as Reflect, where additional information is also included in the prompt. This explains why the Reflect method tends to under-perform in WebShop compared to other methods.

In several cases, FS-CoT-SC can improve the achieved returns of the LLM, especially in GSM8K. However, this comes with the extra cost of needing to prompt the LLM several times (5 in the presented experiments) to perform the SC action selection. In tasks such as HumanEval, where the answer contains long textual answers and potentially several answers can yield the correct outcome, we found that SC cannot be applied. This happens because the LLM will never generate the same answer as before, and the SC action selector cannot choose the most frequent answer.

\subsection{Evaluating Extrinsic Functions: Fine-Tuning}

By closely observing the results in \Cref{sec:exp_pm}, we can conclude that while LLMs can perform well in achieving returns in a wide range of tasks, there is a large room for improvement towards achieving a 100\%  success rate. 
In this section, we aim to explore how SFT and RLFT can help Pangu-Agent increase the returns it achieves. We propose two different pipelines: a Bootstrap SFT (BSFT) that consists of a multi-turn trajectory generation and SFT and a three-step pipeline consisting of trajectory generation, SFT and RLFT. 
Expert trajectory demonstrations, for performing SFT, are always gathered using the OpenChat-3.5 LLM equipped with the structured reasoning abilities of the Pangu-Agent framework. 
We perform BSFT using the OpenChat-3.5 LLM, while the SFT-RLFT pipeline is applied to the Llama 2-7B LLM.
We consider two distinct evaluation paradigms: fine-tuning a different LLM for each task and fine-tuning an LLM in several tasks (e.g. multi-task fine-tuning).

\subsubsection{One Model per Domain}

\paragraph{BSFT:} In the first experiment, we show a combination of the intrinsic functions and the fine-tuning offered by the Pangu-Agent framework.
We first collect data from a diverse set of prompting methods, specifically ZS-CoT, FS-CoT, FS-CoT-React, and FS-CoT-Reflect. 
After this data collection, we run a rejection sampling step, discarding failed trajectories and only keeping the best-performing ones regarding discounted returns.
An SFT step can then be performed on this dataset to improve the method's performance further. Results for the trained model after a single SFT step can be found in \cref{tab:finetune_single_domain_thoughts} under the column "1-step SFT".
Importantly, the model created after one step of this process is still usable under the framework.
Indeed, we can now feed it back to the aforementioned prompting methods and create a higher-quality dataset to fine-tune further.
We repeat this process two more times on top of the OpenChat-3.5 model, each for four epochs over the training data, and we report the results on a held-out test set of ALFWorld episodes in \cref{tab:finetune_single_domain_thoughts}.

\begin{table}[!ht]
\centering
\begin{tabular}{@{}lccccc@{}}
\toprule
\bf \multirow{2}{*}{\textbf{Tasks}}  & \multicolumn{5}{c}{\textbf{OpenChat-3.5}}  \\ \cmidrule(l){2-6}
& \bf Direct & \bf FS-CoT & \bf 1-step SFT & \bf 2-step SFT  & \bf 3-step SFT   \\ \hline \\ [-2ex]
\bf ALFWorld &   0.04  &  0.27  & 0.45 & 0.68 & \bf 0.82 \\ \hline \\[-2ex]
\end{tabular}
\caption{OpenChat-3.5 on ALFWorld with/without fine-tuning on held-out tasks.}
\label{tab:finetune_single_domain_thoughts}
\end{table}

\Cref{tab:finetune_single_domain_thoughts} shows that after a single round of rejection sampling, we can achieve a strong performance in ALFWorld while keeping the model's ability to generate thoughts before actions.

\paragraph{SFT-RLFT:}

That said, fine-tuning on the full trajectories generated by these intrinsic functions is computationally expensive and quickly reaches a point of diminishing returns.
Instead, we propose the use of RL to reach even higher performance in a variety of tasks. To do so, we first perform a fine-tuning step, but on a modified version of intrinsic functions which build the LLM context in a more space-efficient format (i.e.\ a chat-based format \((\bm{o}_t, \bm{a}_t, \bm{o}_{t+1}, \bm{a}_{t+1}, ...)\)).

In this experiment, we fine-tune three distinct Llama 2-7B models on ALFWorld, BabyAI-GoToObj-v0 and BabyAI-GoToRedBlueBall-v0. 
To do so, we first collect 
successful trajectories on training tasks.
To perform SFT, we remove the generated thoughts and only fine-tune on the actions.
After fine-tuning, we evaluate the checkpoint achieving the best score over 512 trajectories in the training domain and test it on the 50 held-out test episodes.

\begin{table}[!ht]
\centering
\begin{tabular}{@{}lcccccc@{}}
\toprule
\bf \multirow{2}{*}{\textbf{Tasks}}  & \multicolumn{2}{c}{\textbf{OpenChat-3.5}} & \multicolumn{4}{c}{\textbf{Llama 2-7B}} \\ \cmidrule(l){2-7}
& \bf Direct & \bf FS-CoT & \bf Original & \bf SFT & \bf SFT+RL & \bf  RL \\ \hline \\ [-2ex]
\bf ALFWorld &   0.04  &  0.27  & 0 & 0.5 & \bf  0.88 & 0.04 \\
\bf BabyAI-GoToObj-v0 &  0.31      &   0.61     &  0.28 & 0.75 & \bf  0.91 & 0.87 \\
\bf BabyAI-GoToRedBlueBall-v0 &  0.11     &  0.43     & 0.04 & 0.21 & \bf 0.77 & 0.69 \\ \hline \\ [-2ex]
\end{tabular}
\caption{Benchmark of OpenChat and Llama 2-7B with/without fine-tuning on held-out tasks.}
\label{tab:finetune_single_domain}
\end{table}

\Cref{tab:finetune_single_domain} shows that fine-tuning first with SFT on successful demonstrations, followed by RL, leads to the largest improvement in success rates.
For complex domains like ALFWorld, it also shows that the SFT step and the intrinsic function (FS-CoT) for trajectory generation are crucial.
This shows the importance of our Pangu-Agent framework where we can benefit from intrinsic functions and fine-tuning. 

\subsubsection{Cross-Domain Model}

In this experiment, we collected additional trajectories on more diverse BabyAI tasks with the same methodology and trained a single cross-domain model on ALFWorld and BabyAI.
Note that no successful trajectories could be generated in BabyAI-UnlockPickup-v0 in the allotted time.
We also remove the thoughts from the successful trajectories to speed up the training.

\begin{table}[!ht]
    \centering
    \begin{tabular}{lcccc}
    \hline \\ [-2ex]
        \bf Task & \bf Training weight & \bf Original & \bf SFT & \bf SFT+RL  \\ \hline  \\ [-2ex]
        \bf ALFWorld & 0.5 & 0 & 0.42 & 0.82 \\ 
        \bf BabyAI-GoToObj-v0 & 0.05 & 0.24 & 0.89 & 0.93 \\ 
        \bf BabyAI-GoToRedBlueBall-v0 & 0.05 & 0.18 & 0.77 & 0.83 \\ 
        \bf BabyAI-GoToRedBallGrey-v0 & 0.05 & 0.08 & 0.72 & 0.81 \\
        \bf BabyAI-GoToLocalS8N7-v0 & 0.05 & 0.22 & 0.8 & 0.87 \\ 
        \bf BabyAI-PickupLoc-v0 & 0.05 & 0.15 & 0.57 & 0.63 \\
        \bf BabyAI-PickupDist-v0 & 0.05 & 0.07 & 0.69 & 0.78 \\
        \bf BabyAI-GoToDoor-v0 & 0.05 & 0.2 & 0.89 & 0.98 \\ 
        \bf BabyAI-OpenRedDoor-v0 & 0.05 & 0.01 & 0.63 & 0.88 \\
        \bf BabyAI-PutNextS7N4Carrying-v0 & 0.05 & 0.07 & 0.39 & 0.66 \\
        \bf BabyAI-UnlockPickup-v0 & 0.05 & 0 & 0 & 0 \\ \hline \\ [-2ex]
        \bf BabyAI-GoToObjDoor-v0 & 0 & 0.13 & 0.82 & 0.81 \\ 
        \bf BabyAI-GoToRedBallNoDists-v0 & 0 & 0.36 & 0.84 & 0.87 \\ 
        \bf BabyAI-GoToOpen-v0 & 0 & 0 & 0.15 & 0.13 \\ 
        \bf BabyAI-GoToObjMazeS4R2-v0 & 0 & 0.05 & 0.23 & 0.24 \\
        \bf BabyAI-Open-v0 & 0 & 0 & 0.25 & 0.51 \\ 
        \bf BabyAI-Unlock-v0 & 0 & 0 & 0.11 & 0.17 \\ 
        \bf BabyAI-PutNextLocal-v0 & 0 & 0 & 0.01 & 0.01 \\ 
        \bf BabyAI-ActionObjDoor-v0 & 0 & 0.05 & 0.64 & 0.78 \\ \hline  \\ [-2ex]
    \end{tabular}
    \caption{Benchmark of Llama 2-7B with/without cross-domain fine-tuning on held-out tasks.}
    \label{tab:finetuning_cross_domain}
\end{table}

\Cref{tab:finetuning_cross_domain} presents the achieved returns of the SFT and RLFT using the Llama 2-7B LLM on ALFWorld and BabyAI tasks.
This experiment establishes that it is possible to successfully SFT and RLFT Llama 2 on multitask training.
The performance in ALFWorld is very close to that achieved by fine-tuning exclusively on ALFWorld.
However, for BabyAI tasks, multi-task training has a clear benefit, achieving even better performance than the specialised model of \Cref{tab:finetune_single_domain}.
It is also capable of generalising to BabyAI tasks that are unseen in training.

\section{Conclusion \& Future Work}

This work presents the Pangu-Agent framework to facilitate future research towards developing generalist AI agents. 
Pangu-Agent builds upon LLMs to address reasoning and decision problems, which allows utilising human priors.
First, we propose a general RL-based objective to optimise the agent's intrinsic and extrinsic functions.
We implemented several intrinsic functions and showcased the modular functionality of our agent and how it can support recent advances in LLM research.
We extensively evaluated Pangu-Agent in several single-agent and multi-agent tasks for different prompt engineering methods across various LLMs, and offered insights about their relative advantages and disadvantages. 

Moreover, we discuss how this framework can fine-tune LLMs through an SFT and RL pipeline. Our results indicate that fine-tuning can improve the agent's performance up to threefold in specific multi-step decision-making domains such as ALFWorld and Baby-AI. We also show how LLMs have the capacity for cross-domain learning by fine-tuning a single LLM and evaluating it simultaneously in multiple domains. 
Finally, we conclude this work by presenting the existing limitations of the current version of Pangu-Agent as well as our vision towards the development of a generalist agent.

\paragraph{Full differentiability:} This work focused on independently optimising the intrinsic and extrinsic functions. 
Looking forward, we envision that Pangu-Agent development will gradually shift towards structured and end-to-end fine-tuning of the framework. This will enable passing gradients between various intrinsic and extrinsic functions, allowing for a more adaptable system.

\paragraph{Real-world applications:}
At present, the performance of Pangu-Agent is evaluated on a small number of single-agent and multi-agent tasks.  
We plan to incorporate more diverse and complex evaluation tasks in future revisions to make Pangu-Agent effective in real-world applications and address the simulation-to-reality gap.
The long-term goal of this work is the development of generalist agents that can assist humans in their everyday tasks and function autonomously in many real-world challenges.

\paragraph{Memory retrieval:}
Another avenue of future research in the Pangu-Agent framework lies in the memory retrieval methods.
The current version of Pangu-Agent supports a long-term memory that stores any information available to each agent, such as its observations, thoughts and actions.
In the future, we aim to incorporate more sophisticated memory retrieval methods, such as embedding similarity from vector databases to allow the agent to incorporate relevant memories in its context window, enabling it to solve the task.

\paragraph{Planning:}
Currently, for planning, we only focus on reasoning tasks.
We intend to integrate and test tree search algorithms in agent-based tasks within interactive environments.
Additionally, we are committed to developing and implementing strategies for efficient long-term planning.
We aim to enhance the planning capabilities of Pangu-Agent, thereby equipping it to tackle real-world challenges and adapt to dynamic environments. 

\paragraph{Tool use:}
Lastly, a significant part of our future Pangu-Agent roadmap is to facilitate integration with external tools.
Pangu-Agent includes a code interpreter for executing simple Python scripts in its current configuration.
However, future versions of Pangu-Agent will support compatibility with various external tools like web search engines, calculators (for instance Wolfram Alpha), and maps.
This expansion will enable a broader deployment of Pangu-Agent across various applications and enable generalisation to tasks beyond its initial learning distribution.

{
\small
\bibliographystyle{IEEEtranN}
\bibliography{reference}
}

\clearpage
\appendix

\section*{Acknowledgements}

The authors would like to express their sincere gratitude to all who contributed to the realisation of this study. A partnership between the team members sustained the project. Jun Wang conceptualised the core research thesis and guided the investigation.
Jianye Hao guided the project from the technical and application aspects. 
Kun Shao assumed the role of project manager, leading the project and making detailed plans for each division.
Haitham Bou-Ammar technically supervised the fine-tuning pipeline and aided in the paper writing, offering insightful recommendations that enhanced the study.
Filippos Christianos and Georgios Papoudakis were instrumental in architecting the proposed agent framework. Their contributions were critical in authoring the principal manuscript, developing the framework, and overseeing the experiments. 
Matthieu Zimmer and James Doran were dedicated to fine-tuning methodologies and the RL pipeline.
Thomas Coste, Zhihao Wu, Jingxuan Chen, and Khyati Khandelwal assisted with the development of the framework, the implementation of various intrinsic methods, and conducted experiments utilising open-source LLMs.
Xidong Feng's efforts focused on implementing tree-search-based methods and performing experiments related to the GPT series.
Jiacheng Liu executed a comprehensive literature review on frameworks for AI agents. 
Zheng Xiong and Yicheng Luo provided support in evaluating the framework.
The authors thank Christopher Mower, Yunfeng Shao, Qiang Ye, Meng Fang, Alexandre Maraval, Antoine Grosnit, Jonas Gonzalez, Rasul Tutunov, and Nan Zhou for their valuable discussions.

\section{Decision-Making and Reasoning Methods}

\begin{figure}[t!]
    \centering
    \includegraphics[trim={3em 4em 3em 3em },clip,width=\textwidth]{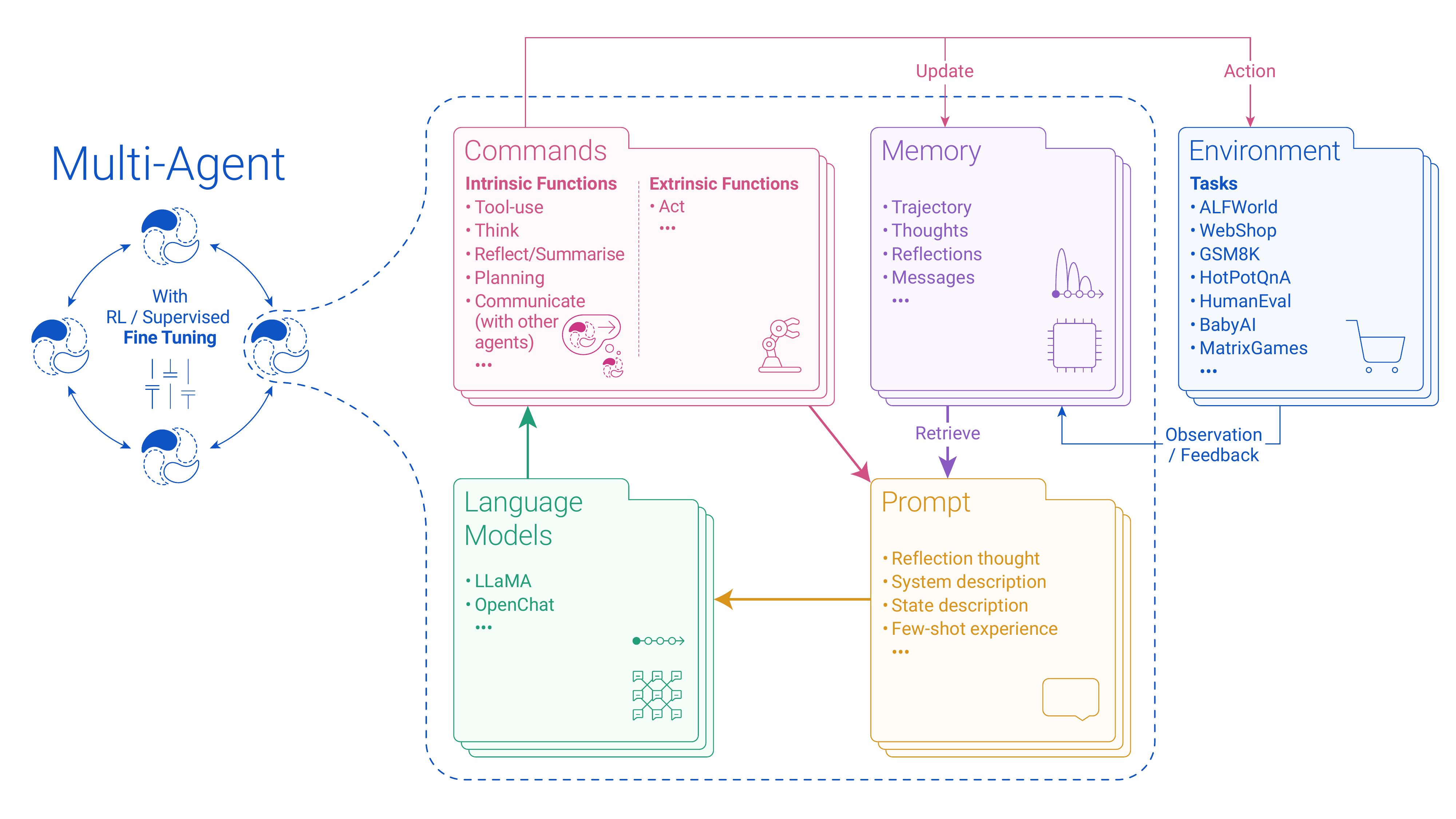}
    \caption{The figure above presents a pictorial depiction of the main components of our proposed agent. On the far left, we can set up a multi-agent environment where each agent can interact with the environment and communicate with other agents. Each agent can be fine-tuned via reinforcement or supervised learning. Each of those agents can support any nesting of intrinsic functions $\Vec{\bm{\mu}}(\cdot)$, such as tool usage, thinking processes, reflecting, planning and others. Those operate on the memory component before producing an action to the outer world via extrinsic processes. Our agent also allows for different prompting strategies and open-source language models, further enabling rigorous experimentation protocols, see \cref{sec:evaluation}.}
    \label{fig:pangu_visual}
\end{figure}

In \cref{sec:evaluation}, we evaluated a variety of both first-order and composite methods across different LLMs. Here, we provide further details on these methods and how they function, with brief summaries in \cref{tab:methods_summary}. We also explain additional features which can be used within the framework, such as agent communication and tool use. A high-level diagram of the Pangu-Agent framework is presented in \Cref{fig:pangu_visual}.

Several of the methods below leverage in-context learning.
In-context learning refers to the collection of methods that utilise analogy-based examples \citep{dong2022survey} to increase the reasoning abilities of LLMs. 
Practically, a number of similar tasks along with their solutions are included in the prompt of the LLM along with other instructions.
Note that in-context learning is not a training paradigm as the pre-trained parameters of the model are not altered. 
However, by conditioning the text generation on the in-context examples, the internal activations of the LLM are shaped towards the task at hand, promoting the emergence of reasoning capabilities. 
This is a significant advantage compared to fine-tuning, which allows to specialise an LLM to a task without the need to fine-tune its parameters, which can be difficult with regards both to computational and data availability considerations.

\begin{table}[h]
\resizebox{\columnwidth}{!}{\begin{tabular}{@{}ll@{}}
\toprule
\textbf{Method}       & \textbf{Description}                                                                                 \\ \midrule
\textbf{Direct}       & Directly prompt the LLM for answers (zero-shot).                                            \\
\textbf{ZS-CoT}       & Zero-Shot, Chain of Thought: Asks the model to think step by step.                          \\
\textbf{FS}           & Few-Shot: In-context examples in which the answer is given directly.                        \\
\textbf{FS-CoT}       & Few-Shot, Chain of Thought: In-context examples with step-by-step thoughts.                 \\
\textbf{FS-CoT-SC}    & Few-Shot, Chain of Thought, Self-Consistency: FS-CoT and checks for consistency in answers. \\
\textbf{FS-CoT-React} & Few-Shot, Chain of Thought, React: Think-then-act React mechanism with FS-CoT examples.   \\ 
\textbf{FS-CoT-Reflect} & Few-Shot, Chain of Thought, Reflect: FS-CoT with an additional reflection step.         \\ 
\textbf{FS-CoT-SwiftSage} & Few-Shot, Chain of Thought, SwiftSage: SwiftSage switching and action buffer with FS-CoT.   \\ 
\textbf{FS-Least-to-Most} & Few-Shot, Least-to-Most: Task decomposition and subgoal answering with FS examples             \\ \bottomrule
\end{tabular}}
\caption{Decision-making and reasoning method summary}
\vspace{-2ex}
\label{tab:methods_summary}
\end{table}

\subsection{First-order Methods}
\label{appdx:intrinsic-extrinsic}

The initial sequence of tokens used as input to the LLM is called the \textit{prompt}.
LLMs, as auto-regressive models, condition the generation of each token on the previous tokens. Therefore, the prompt provided as input to the LLM significantly affects its subsequently generated text. 
Below we present the main prompting techniques that were used in the evaluation of Pangu-Agent.

\textbf{Direct Prompting (Direct):} Direct, or zero-shot, prompting is the simplest way to prompt an LLM. Usually, only a task description and the question at hand are provided. Under direct prompting, no in-context examples are provided to the LLM. It is a simplified variant of the few-shot prompting technique described below, where it is assumed that zero in-context examples are provided.

\textbf{Few-Shot Prompting (FS):} In FS, several question-answer pairs are added to the prompt. Only the task/question and the answer are added to the prompt and not the intermediate steps or thoughts towards the answer. FS has been shown to significantly improve the performance of LLMs in various downstream tasks without requiring any fine-tuning of their parameters \citep{brown2020language}.

\textbf{Few-Shot Chain-of-Thought (FS-CoT) Prompting \citep{wei2022chain}:} CoT refers to step-by-step reasoning through thought generation, eventually leading to the final answer. Several in-context examples are added to the prompt to enable the LLM to develop such reasoning ability. In contrast to FS alone, the question-answer pairs added to the prompt contain intermediate reasoning steps. The prompt is usually augmented with the phrase "think step-by-step". 
This enables the LLM to follow a similar reasoning path when the actual question is provided. 

\textbf{Zero-Shot Chain-of-Thought Prompting (ZS-CoT)\citep{NEURIPS2022_8bb0d291}:} This technique is used to tap into a model's reasoning capabilities by appending the question prompt with "Let's think step-by-step". No other context is provided to the agent, and usually, this technique is indifferent towards the task at hand while also invoking multi-hop reasoning across eclectic tasks with a single template.

\subsection{Composite Methods}\label{appdx:composite}
The first-order methods in the previous part can be combined with more advanced techniques, covered below. In particular, we define: `FS-' which refers to any variations which include context examples before prompting the model for an answer (see Few-Shot Prompting) and `CoT-' which elicits thoughts from the model either through examples when combined with FS (FS-CoT) or simply by asking it to think step by step (ZS-CoT). We use the term \textit{composite} methods to refer to the more advanced methods, which can use first-order prompting techniques and consist of multi-step prompting. 

\textbf{Self-Consistency \citep{wang2022self}:} Self-Consistency (SC) works by repeatedly asking the LLM the same question and expecting the LLM to generate various reasoning paths. Then, the different answers are checked, and the most consistent answer is chosen as the final one. The main strategy for deciding the most consistent answer is majority voting, which chooses the most popular answer.

\textbf{React \citep{yao2022react}:} React is a two-step prompting mechanism that helps to decompose LLM decision-making into reasoning traces and concrete actions. The LLM is first prompted to return a reasoning trace, or \textit{thought}, relevant to solving the task in the current situation. When the LLM is prompted a second time, the reasoning trace is appended to the prompt, and the LLM is asked to return a task-specific action. The reasoning trace provides useful information which can help with commonsense reasoning, progress tracking, planning, and adaptability, to name a few. 

\textbf{Reflect:} In Reflect, the agent is prompted to reflect on its past actions to provide a plan to identify its mistakes and perform better in the upcoming time steps. Hence, the agent is provided with linguistic feedback from the LLM itself to improve its actions. This approach is adapted from \citep{shinn2023reflexion}. Our attempt deviates from the work in \citep{shinn2023reflexion} as we do not maintain a memory across episodes but only provide the agent with the memory from previous time steps in an episode and the most recent reflection. We also reflect at every step instead of reflecting only after three incorrect attempts. Finally, we extend this method to tasks not in the original paper and thus adapt the reflection prompt for these tasks. By default, reflect does not work for single-step tasks with a simple question-answer format. Hence, for such tasks, we introduce a zero-step version of Reflect, which prompts the LLM for a temporary answer and then asks it to reflect on its initial answer before giving a final response. 

\textbf{SwiftSage \citep{lin2023swiftsage}:} In this method, two modules are used: Swift, backed by a smaller language model for quick and intuitive actions, and Sage, backed by a more powerful model for deliberated actions.
The original implementation of this method uses various conditions when switching from Swift to Sage. Deviating from the initial work, our framework uses Swift until five consecutive time steps receive a reward of 0. Sage is prompted to create a plan and provide an action buffer when this occurs. The action buffer is a short list of consecutive actions it believes are the most promising. Actions are selected in order from the action buffer until it is exhausted, at which point we revert to the Swift module. If an action buffer step is invalid, it is skipped, and the next one is considered. This method takes place over multiple time steps and is only relevant for multi-step environments.

\textbf{Least-to-Most \citep{zhou2023leasttomost}:} Similar to ReAct, Least-to-Most is another two-step prompting approach, which asks the LLM first to generate reasoning traces and then produce actions based on the traces. The difference is that, instead of thoughts, the LLM is first prompted to decompose the question into several sub-questions. In the second prompt, the LLM is asked to answer all sub-questions and give a final answer to the original question. Due to context length considerations and our choice of implementation, we deviate from \citep{zhou2023leasttomost} in that all sub-questions are answered in one prompt, and as such, this method only applies to single-step tasks.

\subsection{Additional functions}

\textbf{Communication:} An essential feature of any AI agent is the ability to communicate with other agents, both human and artificial. This enables participation in multi-agent tasks involving several co-existing agents. LLMs appear to be an ideal tool to enable communication between agents. They are grounded in human language, which allows for communication with humans and artificial agents using a standard and universal communication protocol. Pangu-Agent supports pairwise communication between agents, allowing them to participate in multi-agent tasks.

\textbf{Tool Use:} 
Augmenting AI agents with the ability to use external tools can further improve their capacity in many tasks that are hard to solve with LLMs alone, such as mathematical computation and answering questions about current events. 
Pangu-Agent supports tool use through prompting with zero-shot tool descriptions or few-shot tool use examples, which help the agent understand when a tool should be used and how to call it in the correct format. 
The tool output can be either integrated into the agent's current thought or stored in the agent's memory for future use. 
An important example of tool use is a code interpreter \citep{openai2023gpt4,xie2023openagents,xagent2023}, enabling an AI agent to write, run and debug code. 
Pangu-Agent can automatically improve code writing to solve various complicated tasks, such as automated data science by integrating a Python interpreter as a tool and iteratively interacting with it.

\section{Available Tasks} \label{sec:envs}

This section describes the tasks included in the codebase and used for evaluation. We categorise these tasks into two classes based on the number of participating agents: single-agent and multi-agent.

\subsection{Single-Agent Tasks}

\textbf{GSM8K \citep{cobbe2021training}:}
This dataset consists of 8.5k grade-school mathematical problems, curated to have high-quality and diverse language-based problems. It includes natural language solutions instead of pure mathematical equations for evaluation and prompts the agent to evaluate its reasoning and problem-solving ability. The agent receives a score of 1 for correctly answered questions, and 0 otherwise. Context examples for FS and CoT are created using the training set of GSM8K, while the evaluation is performed in the test set, which consists of 1319 questions.

\textbf{HotpotQA \citep{yang-etal-2018-hotpotqa}:}
This Wikipedia-based dataset has 113k question-answer pairs that challenge agents to parse through content and reason over several steps and supporting documents. It is independent of training data or pre-existing knowledge of the agents, uses additional documents for external information, and allows the agent to perform multiple steps for reasoning and comprehension. 
For instance, it provides a paragraph from Wikipedia about a particular band and asks the model a question about that band, the answer to which can be inferred by looking up information in the given paragraph. This environment tests the agent's language comprehension and reasoning ability. Correct answers are rewarded with a score of 1, while wrong answers get 0 reward.

\textbf{ALFWorld \citep{ALFWorld}:}
ALFWorld has been developed to learn text-based policies from TextWorld, which proposes a learning environment for text-based games \citep{côté2019textworld}, and then execute goals from ALFRED, which is a benchmark for the interpretation of grounded instructions for everyday tasks \citep{shridhar2020alfred}. It is a benchmark for language understanding, reasoning ability and task execution skills. 
The ALFWorld framework aligns text descriptions and commands with physically embodied robotic simulation by describing the environment, such as "You are in a room with a desk, chair and lamp" and asks the agent to find a notebook on a desk. Through various actions the agent performs, such as "Go to desk", it is then supposed to complete its task to be rewarded. A completed task will yield a reward of 1, while an uncompleted task will not be rewarded.

\textbf{WebShop \citep{yao2022webshop}:}
WebShop is a simulated e-commerce website with 1.18 million real-world products and 12,087 crowd-sourced text instructions. \citep{yao2022webshop}
In the WebShop environment, the agent is given text instructions for the product environment and asked to navigate different web pages to find, customise, and purchase an item.
The agent can receive observations in either HTML or a text format that strips most of the metadata in HTML. In response, the agent can search the website (when a search box is available) or choose from a pre-defined set of actions. An episode in the WebShop environment ends when the agent selects the buy action. The purchased product is compared with those chosen by a human demonstrator. The returned reward can range from 0 to 1, depending on the similarity to this ground truth, as discussed in \citet{yao2022webshop}.

\textbf{HumanEval \citep{chen2021evaluating}:} HumanEval is a dataset created to benchmark the code generation abilities of language models. It was hand-written by humans, which is quite interesting since the LLMs are thus unlikely to have seen this data during training. Each problem in the dataset comes with a function signature, docstring, and unit tests. In the original paper, the authors asked the model to generate \textit{k} code samples and considered the task completed if any of the samples were correct. In our setup, we evaluate the agent on a more difficult task, checking the validity of a single response. This is done by evaluating the generated function and running the provided unit tests as assert statements. Functions which run successfully and pass the tests will return a reward of 1, and those that fail will return 0.

\textbf{BabyAI-Text \citep{chevalierboisvert2019babyai,carta2023grounding}:} The BabyAI platform consists of several simulated grid-world environments with instruction-following tasks that have gradually increasing levels of difficulty. The environment consists of different objects (balls, boxes, keys, etc.) in various colours, and the agent can pick up, drop or move around objects. The agent is given a 7x7 grid view and textual language instructions for the next steps at each time step. \citet{carta2023grounding} extended BabyAI to provide a text-based benchmark. The reward received at the end of the episode is if the agent reached the goal, discounted by the number of time steps taken to reach it, and 0 otherwise.

\subsection{Multi-Agent Tasks} \label{appdx:ma_tasks}

\textbf{Iterated Prisoner's Dilemma:} The iterated Prisoner’s Dilemma is a classic game-theory example of a non-zero-sum interaction which allows natural language descriptions of self-motivated behaviour in social dilemmas. In it, two agents simultaneously decide whether to cooperate or defect. In our setup, if both agents cooperate at a given iteration, they each receive a reward of -4 (years prison sentence); if they both defect, they each receive a reward of -6; if one defects and the other cooperates, the reward is 0 and -10 respectively. In the case of mutual defection, the joint outcomes are minimised \citep{brookins2023playing}, while in the case of cooperation, the joint outcomes are maximised. It is a test of altruistic behaviour as choosing to defect leads to a loss for both parties while choosing to cooperate can benefit the opposite player. Our Prisoner's Dilemma task is \textit{iterated}, with the agents playing the game five consecutive times while being aware of their partner's answer at the previous time step.  

\textbf{GSM8K \citep{cobbe2021training}:} We implement a multi-agent version of the GSM8K task by assigning an external agent to act as an `expert mathematician' who may receive mathematics doubts or questions from the agent interacting with the task and is asked to clarify them. Similarly, the interacting agent acts as a `mathematics student' tasked with solving the questions and can seek help from an expert if required.

\textbf{HumanEval \citep{chen2021evaluating}:} We use HumanEval in a multi-agent setting by assigning an external agent as an `expert' responsible for helping `interns' with their coding tasks. Similarly, the principal agent is prompted to be an intern, who can seek help from the expert to improve their answers.

\section{Results on Multi-Agent Tasks}
We demonstrate that the Pangu-Agent framework supports multi-agent~\citep{marl-book} scenarios and tasks by evaluating three such tasks introduced in \cref{appdx:ma_tasks}. \Cref{tab:ma} presents the achieved returns for first-order methods where both agents are of the same model type, while \Cref{tab:ma2} presents results when using different model types. In the latter, GPT-3.5 is always used by the teacher for tasks with a teacher-student setup. 
\begin{table}
\centering
\begin{tabular}{@{}clccc@{}}
\toprule
\multirow{2}{*}{\textbf{LLM}}    & \multicolumn{1}{c}{\multirow{2}{*}{\textbf{Method}}} & \multicolumn{3}{c}{\textbf{Task}}                                \\ \cmidrule(l){3-5} 
                                 & \multicolumn{1}{c}{}                                 & \textbf{HumanEval}  & \textbf{GSM8K}      & \textbf{Prisoner's Dilemma}  \\ \hline \\ [-2ex]
\multirow{4}{*}{\rotatebox[origin=c]{90}{\scriptsize \textbf{GPT-4}}}   & \textbf{Direct}                                      & 68.1 \small{$ \pm \ 2.1 $} & 89.7 \small{$ \pm \ 0.4 $} & -12.0 \small{$ \pm \ 0.0 $}     \\
                                 & \textbf{ZS-CoT}                                      & 64.2 \small{$ \pm \ 1.7 $} & 90.2 \small{$ \pm \ 0.6 $} & -11.8 \small{$ \pm \ 0.2 $} \\
                                 & \textbf{FS}                                          & 66.5 \small{$ \pm \ 3.2 $} & 90.0 \small{$ \pm \ 0.7 $}   & -                    \\
                                 & \textbf{FS-CoT}                                      & 67.5 \small{$ \pm \ 1.6 $} & 90.2 \small{$ \pm \ 0.6 $} & -                    \\ \midrule
\multirow{4}{*}{\rotatebox[origin=c]{90}{\scriptsize \textbf{GPT-3.5}}} & \textbf{Direct}                                      & 28.8 \small{$ \pm \ 2.8 $} & 52.2 \small{$ \pm \ 0.7 $} & -9.6 \small{$ \pm \ 0.5 $} \\
                                 & \textbf{ZS-CoT}                                      & 30.5 \small{$ \pm \ 2.6 $} & 59.8 \small{$ \pm \ 0.7 $} & -10.3 \small{$ \pm \ 0.3 $} \\
                                 & \textbf{FS}                                          & 36.6 \small{$ \pm \ 1.7 $} & 53.0 \small{$ \pm \ 0.4 $}   & -                    \\
                                 & \textbf{FS-CoT}                                      & 31.3 \small{$ \pm \ 1.3 $} & 58.7 \small{$ \pm \ 0.3 $} & -          
\\ \midrule

                     \multirow{4}{*}{\rotatebox[origin=c]{90}{\scriptsize \textbf{Llama 2-7B}}} & \textbf{Direct} & 7.0 \small{$ \pm \ 0.3 $}   & 25.0 \small{$ \pm \ 0.5 $}   & -12.0 \small{$ \pm \ 0.0 $} \\
                                & \textbf{ZS-CoT} & 7.7 \small{$ \pm \ 0.8 $} & 24.0 \small{$ \pm \ 0.5 $}   & -12.0 \small{$ \pm \ 0.0 $} \\
                                & \textbf{FS}     & 7.3 \small{$ \pm \ 1.6 $} & 25.7 \small{$ \pm \ 0.9 $} & -                \\
                                & \textbf{FS-CoT} & 5.8 \small{$ \pm \ 0.6 $} & 25.2 \small{$ \pm \ 1.0 $}            & -
                                 \\ \midrule

                     \multirow{4}{*}{\rotatebox[origin=c]{90}{\scriptsize\textbf{OpenChat-3.5}}} & \textbf{Direct} & 23.6 \small{$ \pm \ 1.8$}   & 55.1 \small{$ \pm \ 1.0$}  & -12.0 \small{$ \pm \ 0.0$}\\
                                & \textbf{ZS-CoT} & 19.9 \small{$ \pm \ 2.7$} &  58.7 \small{$ \pm \ 0.6$}  & -12.0 \small{$ \pm \ 0.0$} \\
                                & \textbf{FS}     & 24.2 \small{$ \pm \ 4.4$} & 55.8 \small{$ \pm \ 1.1$} & -                \\
                                & \textbf{FS-CoT} & 22.8 \small{$ \pm \ 3.2$} &  58.3 \small{$ \pm \ 1.1$}           & -    \\ \bottomrule
\end{tabular}
\vspace{1ex}
\caption{Average achieved returns and the standard deviation across three runs, for four first-order prompt engineering methods on three different multi-agent tasks, using four pairs of same LLM type.}
\label{tab:ma}
\end{table}

We observe that using two agents to solve questions in GSM8K and HumanEval reduces returns compared to those presented in \Cref{tab:results:first-order}. This is attributed to the accumulation of error during the multi-agent communication stage. With an extra communication step, the likelihood of the LLM making a mistake or getting confused is increased. 

The iterated Prisoner's Dilemma results show that the framework is capable of supporting multiple agents interacting with the environment at the same time, as is required by this task. The reward shown is the average across the five iterated Prisoner's Dilemma rounds and between both agents. Most LLM types seem to tend towards a reward of -12, or at least below -10. This indicates that the agents frequently, or always for some, choose to defect, which is the expected behaviour of any reasonable agent when the number of iterations is finite and known \citep{pd_exps}.

\begin{table}[]
\centering
\begin{tabular}{@{}cclccc@{}}
\toprule
\multirow{2}{*}{\textbf{\begin{tabular}[c]{@{}c@{}}LLM 1\end{tabular}}} & \multirow{2}{*}{\textbf{\begin{tabular}[c]{@{}c@{}}LLM 2\end{tabular}}} & \multicolumn{1}{c}{\multirow{2}{*}{\textbf{Method}}} & \multicolumn{3}{c}{\textbf{Task}} \\ \cmidrule(l){4-6} 
 &  & \multicolumn{1}{c}{} & \textbf{HumanEval} & \textbf{GSM8K} & \textbf{Prisoner's Dilemma} \\ \midrule
\multirow{4}{*}{\rotatebox[origin=c]{90}{\scriptsize \textbf{GPT-3.5}}} & \multirow{4}{*}{\rotatebox[origin=c]{90}{\scriptsize \textbf{OpenChat-3.2}}} & \textbf{Direct} & 4.3 \small{$ \pm \ 0.5$} & 27.2 \small{$ \pm \ 1.2$} & -10.2 \small{$ \pm \ 0.1$} \\
 &  & \textbf{ZS-CoT} & 3.9 \small{$ \pm \ 0.6$} & 35.2 \small{$ \pm \ 0.7$} & -10.0 \small{$ \pm \ 0.2$} \\
 &  & \textbf{FS} & 3.9 \small{$ \pm \ 0.8$} & 30.3 \small{$ \pm \ 0.4$} & - \\
 &  & \textbf{FS-CoT} & 1.2 \small{$ \pm \ 1.0$} & 
35.9 \small{$ \pm \ 1.4$} & - \\ \midrule
\multirow{4}{*}{\rotatebox[origin=c]{90}{\scriptsize \textbf{GPT-3.5}}} & \multirow{4}{*}{\rotatebox[origin=c]{90}{\scriptsize \textbf{Llama 2-7B}}} & \textbf{Direct} & 10.6 \small{$ \pm \ 3.5$} & 46.5 \small{$ \pm \ 0.2$} & -10.9 \small{$ \pm \ 0.3$} \\
 &  & \textbf{ZS-CoT} & 9.8 \small{$ \pm \ 1.6$} & 55.4 \small{$ \pm \ 0.9$} & -10.7 \small{$ \pm \ 0.0$} \\
 &  & \textbf{FS} & 11.8 \small{$ \pm \ 1.6$} & 46.3 \small{$ \pm \ 1.1$} & - \\
 &  & \textbf{FS-CoT} & 11.0 \small{$ \pm \ 2.0$} & 56.1 \small{$ \pm \ 1.4$} & -\\ \midrule
\multirow{4}{*}{\rotatebox[origin=c]{90}{\scriptsize \textbf{GPT-3.5}}} & \multirow{4}{*}{\rotatebox[origin=c]{90}{\scriptsize \textbf{OpenChat-3.5}}} & \textbf{Direct} & 33.6 \small{$ \pm \ 1.7$} & 53.5 \small{$ \pm \ 0.5$} & -10.8 \small{$ \pm \ 0.1$} \\
 &  & \textbf{ZS-CoT} & 34.2 \small{$ \pm \ 5.4$} & 59.9 \small{$ \pm \ 1.3$} & -10.8 \small{$ \pm \ 0.1$} \\
 &  & \textbf{FS} & 29.6 \small{$ \pm \ 3.4$} & 55.7 \small{$ \pm \ 0.8$} & - \\
 &  & \textbf{FS-CoT} & 31.7 \small{$ \pm \ 2.2$} & 61.5 \small{$ \pm \ 0.7$} & - \\ \bottomrule
\end{tabular}
\vspace{1ex}
\caption{Average achieved returns and the standard deviation across three runs, for four first-order prompt engineering methods on three different multi-agent tasks, using three pairs of different LLM types.}
\label{tab:ma2}
\end{table}

\section{Intrinsic and Composite Functions in the Framework} \label{appdx:nesting_examples}
In this section, we illustrate how simple it is to create, use, and nest intrinsic and composite functions in Pangu-Agent.

\subsection*{Creating new functions from scratch}
New functions can be easily created by defining a new method which defines their behaviour, as follows:
\begin{lstlisting}[language=Python, caption=An example method defining a new intrinsic function.]
class ExampleIntrinsicFunction(Command):
    name: str = "example_intrinsic_function"
    description: str = "An example intrinsic function."
    
    # actions can be defined here, 
    # such as calling the LLM 
    # or storing information in memory
    ...
\end{lstlisting}

\subsection*{Using simple first-order functions}
Using an existing function is as simple as writing a few lines of configuration. An example is given for direct prompting, which only uses the Act function:

\begin{lstlisting}[language=yaml, caption=A configuration defining the Direct prompting method.]
main_flow:
  _target_: pangu.commands.SequentialFlow
  sequence:
    - _target_: pangu.commands.Act
prompt_builder:
  default_kwargs:
    cot_type: zero_shot
\end{lstlisting}

\subsection*{Adding Chain of Thought}
Chain of Thought (CoT) is controlled by a single variable, such that including CoT in our agent is as easy as changing the \texttt{cot\_type} configuration variable value. Direct prompting can thus be transformed into ZS-CoT:

\begin{lstlisting}[language=yaml, caption=A configuration defining the ZS-CoT prompting method.]
main_flow:
  _target_: pangu.commands.SequentialFlow
  sequence:
    - _target_: pangu.commands.Act
prompt_builder:
  default_kwargs:
    cot_type: zs-cot                      # change the CoT type
\end{lstlisting}

\subsection*{Using composite methods}
Composite functions can be used within the agent by nesting functions, such as Reflect before Act. This is as simple as adding this command to the configuration:

\begin{lstlisting}[language=yaml, caption=A configuration defining the Reflect method.]
main_flow:
  _target_: pangu.commands.SequentialFlow
  sequence:
    - _target_: pangu.commands.Reflect    # add Reflect
    - _target_: pangu.commands.Act
...
\end{lstlisting}

\subsection*{Nesting functions}
It is even possible to further nest intrinsic functions together, such as Reflect and Think:

\begin{lstlisting}[language=yaml, caption=A configuration defining a combination of the Reflect and React methods.]
main_flow:
  _target_: pangu.commands.SequentialFlow
  sequence:
    - _target_: pangu.commands.Reflect
    - _target_: pangu.commands.Think      # add Think after Reflect
    - _target_: pangu.commands.Act
...
\end{lstlisting}

\subsection*{Creating new composite functions from existing functions}

In order to create new composite functions from existing functions, one can simply define a new function putting the building blocks together. An example of doing this for the previous example of nested functions follows:

\begin{lstlisting}[language=Python, caption=An example method defining a new composite function.]
CompositeFunction = partial(
  SequentialFlow,
  name="composite_function",
  description="An example composite function.",
  sequence=[Reflect(),Think(),ConsiderAction(),ExecutePlannedAction()]
)
\end{lstlisting}

The new composite function can then simply be used by adding it to the configuration as before.

\subsection*{Letting the agent choose methods}
We can even let the agent decide which of many functions it wants to use by using a DecisionFlow in the configuration:

\begin{lstlisting}[language=yaml, caption=A configuration which lets the agent choose between two methods.]
main_flow:
  _target_: pangu.commands.DecisionFlow          # let model choose 
  choices:
    - _target_: pangu.commands.CompositeFunction # new function
    - _target_: pangu.commands.Act
...
\end{lstlisting}

The DecisionFlow presents the list of specified methods to the LLM and the LLM is instructed to choose a method it believes will help it most. Thus, at every time step, the LLM is able to decide and change according to which method it will operate. Decision points and sequences of intrinsic actions can be nested to create complex reasoning steps for the AI agent.

\subsection*{Implementing composite methods}
Using the procedures described above, we can define any composite method. For example, for the Self-Consistency method: 

\begin{lstlisting}[language=Python, caption=A method definition for Self-Consistency.]
SelfConsistencyAct = partial(
  SequentialFlow,
  name="self_consistency_act",
  description="Run CoT multiple times and select the most consistent answer.",
  sequence=[
    ConsiderAction(),
    ConsiderAction(),
    ConsiderAction(), # as many times as needed  
    ConsistencyOnDiverseActions(),ExecutePlannedAction()
  ]
)
\end{lstlisting}
First, $n$ instances of \texttt{ConsiderAction} are called to generate $n$ answers to a same prompt. Then \texttt{ConsistencyOnDiverseActions} is used to select the most consistent answer. Finally, the answer is executed within the environment using \texttt{ExecutePlannedAction}.

We also show how we can define a method resembling ReAct by defining a configuration file, using our framework. In the ReAct method, the language model is optionally asked to perform a distinct thinking step before returning an answer. We can implement this using a \texttt{DecisionFlow} as follows:
\begin{lstlisting}[language=yaml, caption=A configuration defining our version of ReAct.]
main_flow:
  _target_: pangu.commands.DecisionFlow
  choices:
    - _target_: pangu.commands.SequentialFlow
      sequence:
        - _target_: pangu.commands.Think
        - _target_: pangu.commands.Act
    - _target_: pangu.commands.Act
\end{lstlisting}
At every time step, the LLM is asked whether it wants to perform a distinct thinking step before acting, or whether it wants to act directly.

\section{Results on Search-enhanced Planning}
\label{apx:planning}
We present the results of planning-enhanced LLM in \cref{tab:results:planning}. We choose two models - GPT-3.5 and task-tuned Llama 2-7B with trained value function (obtained from \citep{feng2023alphazero}) and test them on Game24 \citep{yao2023tree} and GSM8K. Based on the table, we mainly have three conclusions. Firstly, the LLM's basic task-solving ability (indicated by zero-shot, few-shot or few-shot CoT performance) largely determines the performance of different planning methods. Stronger base models (the Llama 2-7B-SFT in Game24 and the GPT-3.5 in GSM8K) lead to better tree-search enhanced generation. Our second and third conclusion aligns with the findings shown in \citep{feng2023alphazero}. Secondly, we find that for small-scale search problems presented in the table, different search algorithms seem to behave similarly. This conclusion aligns with the findings of \citep{feng2023alphazero}. Our final finding is that planning-enhanced generation has relatively limited improvements for GPT-3.5 and we even observe the performance drop. We believe this is because of GPT-3.5's weak evaluation ability and this phenomenon also aligns with other papers studying LLM's limited self-evaluation ability \citep{feng2023alphazero, huang2023large, stechly2023gpt}. When we have a stronger evaluator, such as a task-tuned value function (used in Llama 2-7B-tuned in our setting) or GPT-4 with well-designed prompts (used in ToT \citep{yao2023tree}), we can still obtain performance gain compared to other baselines.
\begin{table}
\centering
\caption{\label{tab:results:planning} Accuracy on the Game24/GSM8K environment for different tree-search methods}
\begin{tabular}{@{}lcccccccc@{}}
\toprule
 \textbf{Env}     &  \textbf{Model} & \textbf{ZS-CoT}  & \textbf{FS}   & \textbf{FS-CoT}  & \textbf{BFS} & \textbf{DFS} & \textbf{MCTS} \\ 
              \midrule
\textbf{Game24} & GPT-3.5 \citep{yao2023tree} &  -  & 5.0 & 1.0 & 10.0 & 3.0 & 5.0 \\
\textbf{Game24} & Llama 2-7B-tuned \cite{feng2023alphazero}&   30.0 & -  & - & 72.0 & 74.0  &  73.0 \\
\midrule
\textbf{GSM8K} & GPT-3.5 \citep{yao2023tree}  & 69.2  & 35.0 & 66.4 & 64.1 & 61.4 &  65.5\\
\textbf{GSM8K} & Llama 2-7B-tuned \cite{feng2023alphazero} & 41.4 & - & - & 54.4  & 53.7 & 53.0  \\
\bottomrule
\end{tabular}
\end{table}

\section{Related Work}
\label{sec:related}
In this appendix, we provide a brief review of relevant prior works about LLM agent frameworks, readers can refer to table 1 for more intuitive comparisons between some well-known LLM agent frameworks.
\subsection{Single-Agent Frameworks}
The rapid advancement in the generalistic capabilities of LLMs has catalysed the development of more complex and capable LLM-based agents, across both text-based and multimodal environments.

A range of proof-of-concept works have demonstrated the reasoning and planning abilities of LLMs. Examples such as Chain-of-Thought~\citep{wei2022chain}, Self Consistency~\citep{wang2022self}, Tree of Thoughts~\citep{yao2023tree,long2023large}, and Graph of Thoughts~\citep{yao2023beyond,besta2023graph} exhibit models' aptitude for structured conceptual exploration. Other efforts emphasise the iterative self-improvement capacities of LLMs, including Self-Refine~\citep{madaan2023self}, ReAct~\citep{yao2022react}, ReWOO~\citep{xu2023rewoo}, Reflexion~\citep{shinn2023reflexion}, and Chain-of-Hindsight~\citep{liu2023languages}. These works provide evidence that LLMs have the basic underlying capacity to support reasoning, planning, memory, and iterative self-improvement needed to form capable agents.

Other efforts have focused on developing LLM agent frameworks to tackle more complex settings. Single-agent frameworks incorporate the LLM as a central controller alongside other modular components.  
For example, the GPT-4\citep{openai2023gpt4} model on the OpenAI platform can serve as a personal assistant agent, leveraging plugins, a code interpreter, and web integration to operate in closed and open environments. Open Interpreter~\citep{openinterpreter} and OpenAgents~\citep{xie2023openagents} are open-source implementations which try to emulate the agent structure on the OpenAI platform. Transformer Agents\citep{wolf2019huggingface} and LangChain~\citep{LangChain} are open-source repositories that have been developed to help developers build a single LLM agent more easily with built-in functionalities. AutoGPT~\citep{gravitas2023auto} focuses on utilising LLMs for achieving flexible goals within a single-agent structure. Other proposed single-agent frameworks such as Gentopia~\citep{xu2023gentopia} and AgentGPT\citep{agentgpt} follow similar ideas. 
In contrast to these, the Pangu-Agent framework, first, proposes and builds upon a novel optimisation objective which offers modularity and flexible design, and second, allows fine-tuning the intrinsic and extrinsic functions to improve the achieved returns of the agent. A comparison between Pangu-Agent and the remaining frameworks is summarised in \cref{tab:agent_summary}.

\subsection{Multi-Agent Frameworks}
To unlock more advanced features which might be required for complex tasks, multi-agent frameworks have been studied. While preliminary works like Camel~\citep{li2023camel}, Multi-Agent Debate~\citep{liang2023encouraging,du2023improving}, BabyAGI~\citep{babyagi}, CHATDEV~\citep{qian2023communicative}, MetaGPT~\citep{hong2023metagpt}, RestGPT\citep{song2023restgpt}, and ProAgent~\citep{zhang2023proagent} demonstrate the potential for multiple LLMs to collaborate on challenges difficult for a single agent, their inter-agent communication patterns are relatively fixed.

Recent work has focused on making multi-agent communication more flexible. BOLAA~\citep{liu2023bolaa} utilises a controller module to manage communication between "labor" agents. AgentVerse~\citep{chen2023agentverse} combines expert recruitment with collaborative decision-making to adapt to different tasks. AutoAgents~\citep{chen2023auto} employs observer agents to enhance recruitment, collaboration, and action execution process. Dylan~\citep{liu2023dynamic} 
introduces a dynamic multi-agent architecture and automatic agent team optimization. Some frameworks further emphasise customisability and extensibility. Configurability is a shared feature among these frameworks. AGENTS~\citep{zhou2023agents} uses configurable symbolic plans called Standard Operating Procedures to control agent behaviours. AutoGen~\citep{wu2023autogen} focuses on customisable conversable agents and conversation programming to enable flexible conversation patterns for collaborative reasoning between agents. 

\end{document}